\documentclass{bytedance_seed}
\usepackage[T1]{fontenc}
\usepackage[utf8]{inputenc}
\usepackage{amsmath,amssymb}
\usepackage{graphicx}
\usepackage{fontawesome5}
\usepackage{booktabs}
\usepackage{multirow,makecell}
\usepackage{tabularx}
\usepackage{enumitem}
\usepackage{needspace}
\usepackage{xcolor}
\usepackage{url}
\usepackage{hyperref}
\graphicspath{{figures/}}

\newcommand{\benchmark}{DeliverBench}

\title{DeliverBench: Benchmarking AI Agents on Delivering Realistic Multi-Stage Professional Projects}
\author[1,2,3,*]{Yumou~Liu}
\author[1,*]{Mingju~Chen}
\author[1,*,\dagger]{Zhoufutu~Wen}
\author[1]{Qianyu~Yang}
\author[1]{Zijie~Wang}
\author[3]{Sihan~Chen}
\author[1]{Yeqian~Liu}
\author[1]{Chenghao~Yang}
\author[1]{Tiliang~Duan}
\author[1]{Qiyu~Cai}
\author[1]{Yanglihong~Xiao}
\author[1]{Jiayun~Dong}
\author[1]{Jianlan~Ma}
\author[5]{Shenhao~Miao}
\author[5]{Yuhan~Cao}
\author[1]{Lu~Xu}
\author[\ddagger]{Que~Shen}
\author[1]{Tian~Gao}
\author[6]{Ziyu~Liang}
\author[7]{Eric~Yang}
\author[4]{Ge~Zhang}
\author[1]{Jianpeng~Jiao}
\author[2,3,\dagger]{Xuanhe~Zhou}
\author[2,3]{Fan~Wu}

\affiliation[1]{Humanlaya}
\affiliation[2]{Theseus Labs}
\affiliation[3]{Shanghai Jiao Tong University}
\affiliation[4]{TokenWave.AI}
\affiliation[5]{GaugeForge}
\affiliation[6]{High Quant Investments}
\affiliation[7]{Gradient}
\contribution{%
  \href{https://deliverbench.github.io/}{\faGlobe\enspace\textbf{Homepage}}%
  \hspace{1.8em}%
  \href{https://github.com/theseus-labs-rsi/deliverbench}{\faGithub\enspace\textbf{GitHub Code}}%
  \hspace{1.8em}%
  \href{https://huggingface.co/datasets/deliverbench/DeliverBench}{\faDatabase\enspace\textbf{Hugging Face Dataset}}%
}
\abstract{LLM agents are judged by what they deliver, and in professional work what gets delivered is a project, not a task.
A project pays off only if the whole package holds together, and one weak part can limit the value of the rest.
Yet existing benchmarks mostly test agents on one task at a time, leaving a gap between the capability we measure and the outcome that makes agents useful.
To address this gap, we introduce \textbf{DeliverBench}, which tests whether agents can carry a project from brief to final delivery.
Reconstructed from real materials with 13 domain experts, its 11 projects span settings such as financial due diligence, legal contract drafting, and drug safety reporting.
They comprise 150 subtasks, with 2--15 subsequent releases of new materials and client feedback per project.
Later subtasks build on the agent's own earlier deliverables, so unresolved errors can carry into downstream work.
Expert rubrics assess each subtask submission and the final delivery package.
Across seven model--harness systems, the best scores 65.0/100 on final delivery and passes 50.0\% of subtasks at a 90-point threshold.
Together, these results suggest that task-level capability does not simply compose into delivery; delivery also depends on keeping the work coherent as it evolves.
}

\begin{document}
\raggedbottom
\maketitle
\begingroup
\renewcommand{\thefootnote}{\ensuremath{*}}
\begin{NoHyper}
\footnotetext{Equal contribution.}
\end{NoHyper}
\endgroup
\begingroup
\renewcommand{\thefootnote}{\ensuremath{\dagger}}
\begin{NoHyper}
\footnotetext{Corresponding authors: Zhoufutu Wen (\texttt{maxwell@humanlaya.com}) and Xuanhe Zhou (\texttt{zhouxuanhe@sjtu.edu.cn}).}
\end{NoHyper}
\endgroup
\begingroup
\renewcommand{\thefootnote}{\ensuremath{\ddagger}}
\begin{NoHyper}
\footnotetext{Independent researcher.}
\end{NoHyper}
\endgroup

\begingroup\setlength{\parskip}{0pt}\noindent\begin{minipage}{\linewidth}
  \centering
  \captionsetup{skip=8.75pt}
  \includegraphics[width=\linewidth]{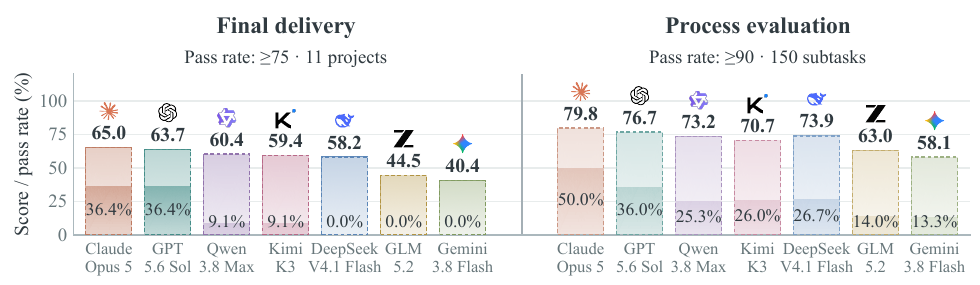}
  \captionof{figure}{\textbf{Main leaderboard on DeliverBench.} Bar tops show mean scores; filled portions show pass rates. Thresholds are 75 for 11 project deliveries and 90 for 150 subtasks per system.}
  \label{fig:main-leaderboard}
\end{minipage}\par\endgroup
\clearpage

\begin{figure}[h]
  \centering
  \includegraphics[width=\linewidth]{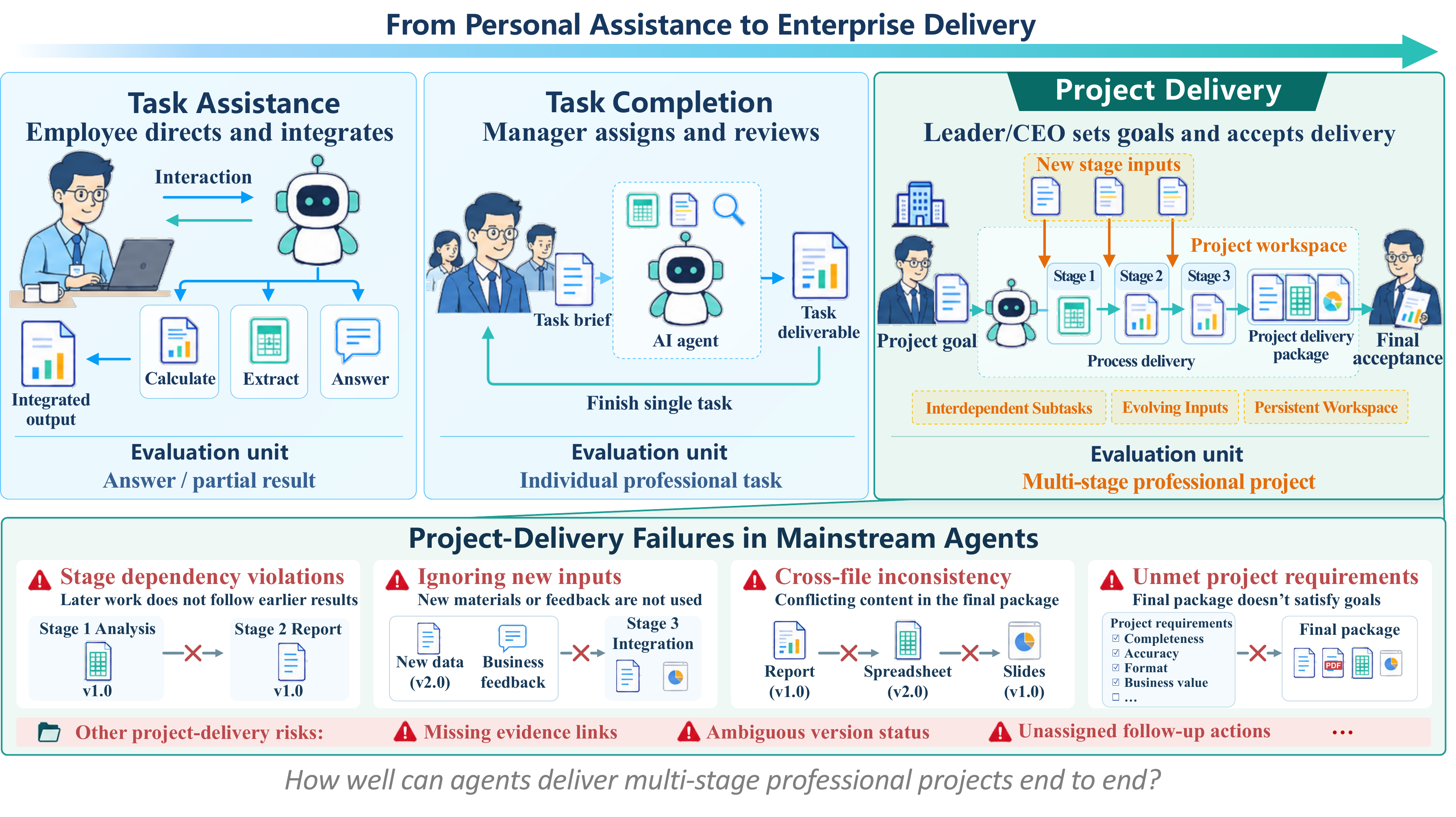}
  \caption{\textbf{Task assistance, task completion, and project delivery.} As delegation expands, the evaluation unit shifts from partial results and individual tasks to a multi-stage project. DeliverBench follows an agent's work through a persistent workspace, new inputs, intermediate handoffs, and final acceptance. The lower panel illustrates delivery failures that require tracking dependencies, revisions, and the joint state of the final package.}
  \label{fig:motivation}
\end{figure}

\section{Introduction}
\label{sec:introduction}

\begin{table}[htbp]
\centering
\caption{Comparison of published benchmark protocols. \textbf{Evaluation unit}: the scope of work in one benchmark instance. \textbf{Agent's artifact handoff}: later component tasks use or revise artifacts generated by the same agent in earlier tasks. \textbf{Project delivery}: an integrated final package is evaluated against a shared project outcome, including consistency and acceptance requirements. $\checkmark$: explicit support; $\triangle$: limited or related support; $\times$: not specified. Appendix~\ref{app:comparison-definitions} gives the operational definitions and evidence. Ratings follow these criteria and are based on each benchmark's published protocol.}
\label{tab:benchmark-comparison}
\begingroup
\small
\setlength{\tabcolsep}{3pt}
\renewcommand{\arraystretch}{1.13}
\resizebox{\textwidth}{!}{%
\begin{tabular}{@{}llcccccc@{}}
\toprule
\textbf{Benchmark} & \textbf{Evaluation unit} & \makecell{\textbf{Agent's artifact}\\\textbf{handoff}} & \makecell{\textbf{Project}\\\textbf{delivery}} & \makecell{Evolving\\inputs/feedback} & \makecell{Long-horizon\\tasks} & \makecell{Explicit\\subtasks} & \makecell{Intermediate\\evaluation} \\
\midrule
GDPval~\citep{patwardhan2025gdpval} & Professional task & $\times$ & $\triangle$ & $\times$ & $\checkmark$ & $\times$ & $\times$ \\
APEX-Agents~\citep{vidgen2026apexagents} & \makecell[l]{Professional task} & $\times$ & $\triangle$ & $\times$ & $\checkmark$ & $\times$ & $\times$ \\
DAYJOB~\citep{finley2026dayjob} & Professional task & $\times$ & $\checkmark$ & $\times$ & $\checkmark$ & $\times$ & $\triangle$ \\
JobBench~\citep{li2026jobbench} & Professional task & $\times$ & $\triangle$ & $\times$ & $\checkmark$ & $\triangle$ & $\triangle$ \\
OfficeBench~\citep{wang2024officebench} & \makecell[l]{Office task} & $\triangle$ & $\triangle$ & $\times$ & $\checkmark$ & $\triangle$ & $\times$ \\
OSWorld 2.0~\citep{yuan2026osworld2} & Computer-use workflow & $\triangle$ & $\checkmark$ & $\checkmark$ & $\checkmark$ & $\triangle$ & $\checkmark$ \\
AutomationBench~\citep{shepard2026automationbench} & Cross-app workflow & $\triangle$ & $\times$ & $\times$ & $\checkmark$ & $\triangle$ & $\triangle$ \\
WorkArena++~\citep{boisvert2024workarenapp} & \makecell[l]{Compositional task} & $\triangle$ & $\times$ & $\times$ & $\checkmark$ & $\checkmark$ & $\checkmark$ \\
TheAgentCompany~\citep{xu2024theagentcompany} & Workplace task & $\triangle$ & $\triangle$ & $\checkmark$ & $\checkmark$ & $\triangle$ & $\checkmark$ \\
PaperBench~\citep{starace2025paperbench} & ML paper replication & $\triangle$ & $\checkmark$ & $\times$ & $\checkmark$ & $\triangle$ & $\triangle$ \\
StartupBench~\citep{zhu2026startupbench} & \makecell[l]{Workflow task} & $\times$ & $\triangle$ & $\times$ & $\checkmark$ & $\times$ & $\times$ \\
Workspace-Bench 1.0~\citep{tang2026workspacebench} & \makecell[l]{Workspace task} & $\times$ & $\triangle$ & $\times$ & $\checkmark$ & $\times$ & $\triangle$ \\
SQBench~\citep{sun2026sqbenchbenchmarkevaluatingtask} & Knowledge-work task & $\times$ & $\triangle$ & $\times$ & $\triangle$ & $\times$ & $\triangle$ \\
\midrule
\textbf{DeliverBench} & \makecell[l]{\textbf{Multi-stage}\\\textbf{professional project}} & $\pmb{\checkmark}$ & $\pmb{\checkmark}$ & $\pmb{\checkmark}$ & $\pmb{\checkmark}$ & $\pmb{\checkmark}$ & $\pmb{\checkmark}$ \\
\bottomrule
\end{tabular}}
\endgroup
\end{table}

As AI agents become capable of producing professional work products, they are increasingly expected to carry entire projects from an initial brief to final handoff~\citep{patwardhan2025gdpval,xu2024theagentcompany}.
Figure~\ref{fig:motivation} contrasts task assistance, task completion, and project delivery. In the last setting, clients define project goals and acceptance requirements, while agents take responsibility for the intervening work.
The value of this delegation depends on whether the resulting outputs jointly serve the client's objective.
Yet most benchmarks still organize evaluation around individual assignments.
Success on those assignments does not establish whether an agent can carry its outputs into later subtasks and revise them as new information arrives.
This raises a central question: \emph{How well can AI agents deliver multi-stage professional projects end to end?}

Professional project delivery provides a demanding setting for this question.
In business-to-business services, clients commission projects whose deliverables jointly support a business purpose, while new evidence and feedback arrive as work unfolds.
Recent benchmarks bring realistic professional tasks, cross-application workflows, and complex workspaces into evaluation~\citep{vidgen2026apexagents,yuan2026osworld2,tang2026workspacebench}; project delivery brings these demands together across successive handoffs.
In a financing project, due diligence findings inform financing terms, which enter a financial model supporting lender coordination.
When client feedback corrects a payment amount, the correction needs to reach the operative schedule used in subsequent calculations, beyond appearing in a change note.
In an audit, an evidence limitation recorded in the summary constrains the conclusions supported by the detailed workpaper.
Effective coordination therefore depends on identifying which downstream calculations, conclusions, and actions a change affects~\citep{malone1994coordination}.
These demands, illustrated in the lower panel of Figure~\ref{fig:motivation}, make project delivery a natural testbed for whether agents can track dependencies, respond to new information, and reconcile related deliverables throughout a project.

We introduce \textbf{\benchmark{}}, which takes a \textit{multi-stage professional project} as its evaluation unit.
Reconstructed from real materials with 13 domain experts, the benchmark comprises 11 projects across seven domains, with 150 interdependent subtasks and 111 required final deliverables (Section~\ref{sec:dataset-quality}).
Experts provide source records, intermediate and final deliverables, feedback, and acceptance standards; we reconstruct the stages, subtask instructions, and material-release schedules with their guidance.
Agents work in a persistent workspace as new materials and client feedback arrive.
Interactive feedback can respond to the agent's actual requests and submitted artifacts within expert-defined roles and available facts.
Later subtasks directly use the agent's own earlier deliverables, without replacing them with reference artifacts.
An unresolved upstream error can consequently enter downstream analysis or persist across several mutually consistent files.
Table~\ref{tab:benchmark-comparison} compares these project-level requirements with existing benchmark protocols.

Scoring each file separately does not establish whether the files agree or jointly meet the project's handoff requirements.
\benchmark{} therefore combines process evaluation with final project delivery evaluation.
Rubrics reviewed by domain experts assess each subtask submission against its requirements and the information available at that point.
The final package is assessed for \textit{artifact quality}, \textit{cross-file consistency}, and \textit{package acceptance}.
Each criterion also carries one primary quality-dimension label: correctness, completeness, consistency, professionalism, or usability.
These views provide both a record of intermediate performance and an assessment of the package handed to the recipient.
For final deliverables produced by GPT-5.6 Sol--Codex on three projects, expert and evaluator judgments agree on 446 of 478 rubric decisions (93.31\%; Section~\ref{sec:evaluation-validation}).

We evaluate seven model--harness systems on \benchmark{} (Figure~\ref{fig:main-leaderboard}).
Claude Opus 5 attains the highest mean final-delivery score (65.0 out of 100), followed by GPT-5.6 Sol (63.7).
Even the strongest system passes only half of the subtasks at the 90-point threshold; the remaining systems pass at most 36.0\%.
Every system scores lower on final delivery than on process evaluation, by 11.3--18.5 points, and 69 of 77 system--project outcomes show the same pattern (Section~\ref{sec:main-results}).
Inspection of the preserved artifacts reveals concrete failures in carrying work forward: a closing memo and task register disagree on the current blocker, and a superseded design decision remains active in the decision register (Section~\ref{sec:case-failures}).
These results show that completing individual subtasks does not by itself ensure a usable final package; project delivery also depends on keeping the work coherent as it evolves.

\Needspace{8\baselineskip}
Our contributions are as follows.
\begin{itemize}[leftmargin=*, topsep=4pt, partopsep=0pt, itemsep=5pt, parsep=0pt]
  \item \textbf{A benchmark of multi-stage professional projects.} We construct \benchmark{} from real project materials with domain experts across seven domains. Its design captures the project lifecycle through shared objectives, interdependent subtasks, evolving information, and the continued use and revision of agent-produced artifacts.
  \item \textbf{An evaluation framework for process and final delivery.} We combine subtask assessment with final-package evaluation, covering artifact quality, cross-file consistency, and package acceptance. Expert-reviewed rubrics, five quality dimensions, and comparisons with expert judgments support detailed assessment and analysis of delivery quality.
  \item \textbf{A systematic empirical study of project delivery.} We evaluate seven model--harness systems across performance, quality, stability, and resource use. Analyses of preserved artifacts and trajectories characterize the process--delivery gap, identify failures in maintaining accumulated work, and examine recovery through upstream artifact repair.
\end{itemize}

\section{Related Work}
\label{sec:related-work}

Table~\ref{tab:benchmark-comparison} compares representative benchmarks along three connected directions: professional work products, dependent workflows, and evidence of successful delivery.

\textbf{Professional and office tasks.}
Professional benchmarks increasingly ask whether agents can produce usable work under the requirements of an occupation.
\textit{GDPval}, \textit{APEX-Agents}, and \textit{Agents' Last Exam (ALE)} ground evaluation in expert-designed tasks and professional outputs~\citep{patwardhan2025gdpval,vidgen2026apexagents,sun2026ale}, making domain judgment and the quality of the resulting artifacts central to assessment.
The supporting workspace also shapes the work: \textit{JobBench} and \textit{DAYJOB} require agents to interpret heterogeneous records, identify relevant evidence, and produce analyses that satisfy detailed professional requirements~\citep{li2026jobbench,finley2026dayjob}.
\textit{OfficeBench} complements this emphasis on professional content by testing the coordination of applications needed to complete office assignments~\citep{wang2024officebench}.
Together, these designs connect task difficulty to the materials, tools, and judgments involved in producing a usable deliverable.
\benchmark{} builds on this perspective by following those work products beyond an individual assignment: a submitted analysis becomes an input to later subtasks.

\textbf{Workflow and project benchmarks.}
Workflow benchmarks examine how agents coordinate actions when later operations depend on earlier results.
\textit{WorkArena++} and \textit{AutomationBench} study this problem through compositional enterprise tasks and policy-guided API workflows~\citep{boisvert2024workarenapp,shepard2026automationbench}, where retrieving information, making decisions, and changing application state must form a coherent sequence.
Other environments make interaction and changing context part of execution: \textit{OSWorld~2.0} includes dynamic task updates, while \textit{TheAgentCompany} embeds tasks in a workplace with simulated colleagues~\citep{yuan2026osworld2,xu2024theagentcompany}.
\textit{$\tau$-bench} and \textit{$\pi$-Bench} further study how requests are resolved through policy-guided conversations and proactive assistance across sessions, respectively~\citep{yao2024taubench,zhang2026pibench}.
These settings test whether agents can maintain task context and choose appropriate actions as work unfolds.

File dependencies and deliverable-oriented workflows bring a related challenge into view.
\textit{Workspace-Bench} emphasizes reasoning across relationships in an existing file workspace, while \textit{StartupBench} grounds complete workflows in practical user demand and evaluates their deliverables~\citep{tang2026workspacebench,zhu2026startupbench}.
\benchmark{} focuses on dependencies across successive submissions by the same agent: later subtasks directly use its earlier artifacts, and new materials or feedback can require revising both an upstream result and the work that depends on it.
Following these dependencies makes error propagation, revision, and repair part of the project being evaluated, alongside the execution of each subtask.

\textbf{Process and delivery evaluation.}
Progress, execution behavior, and final submissions provide different evidence of agent capability.
\textit{AgentBoard} measures incremental progress through state matching and subgoal completion, whereas \textit{Agent-as-a-Judge} evaluates development requirements and their dependencies in DevAI~\citep{ma2024agentboard,zhuge2024agentasjudge}.
For completed work, \textit{PaperBench} combines rubric-based assessment of research submissions with execution in a fresh environment; \textit{SQBench} separates functional completion from risk checks supported by execution evidence~\citep{starace2025paperbench,sun2026sqbenchbenchmarkevaluatingtask}.
These approaches make partial achievement, functional correctness, and execution constraints visible in different ways.
In a professional project, a further question is whether the latest artifacts jointly support handoff, even when earlier submissions satisfied their local requirements.
\benchmark{} therefore preserves and evaluates each subtask submission against the information available at that point, then assesses artifact quality, cross-file consistency, and package acceptance in the final delivery.
This connects evidence of local progress to the accumulated work that downstream tasks and the final recipient actually use.

\Needspace{8\baselineskip}
\section{\benchmark{}}
\label{sec:benchmark}

\benchmark{} evaluates professional project delivery through intermediate submissions and a complete, consistent, and usable final package.

\subsection{Professional Projects as the Evaluation Unit}
\label{sec:problem}

A \emph{professional project delivery instance} specifies a professional role, a shared business objective, initial materials, interdependent subtasks, and a final multi-file package.
A multi-stage project organizes these subtasks around delivery milestones: each stage contains one or more subtasks, and each subtask ends with a submission preserved for process evaluation.
The project concludes with a final delivery package.
Each project combines three elements:
\begin{itemize}[leftmargin=*, itemsep=5pt, parsep=0pt]
  \item \textbf{Shared delivery requirements.} The project objective, scope, required files, and acceptance conditions specify what the recipient needs to review, act on, or continue the work. Deliverables must meet professional standards and jointly satisfy these requirements.
  \item \textbf{Interdependent subtasks in a persistent workspace.} Earlier files remain available as working inputs. Their data, assumptions, and conclusions constrain later calculations, judgments, and formal documents; the agent must check, revise, and integrate them as the project proceeds.
  \item \textbf{New inputs during execution.} Materials and client decisions arrive over time. Interactive events also provide client or reviewer feedback conditioned on the agent's submitted artifacts, requiring updates to earlier work and the deliverables that depend on it.
\end{itemize}
Table~\ref{tab:project-names} maps project identifiers to their commonly used names.
Appendix~\ref{app:representative-workflow} illustrates the setting through Project Mars Refinancing (P02): due diligence informs financing terms, those terms enter a financial model, and the resulting analyses support lender coordination and final handoff.

\begin{table}[htbp]
  \centering\small
  \setlength{\tabcolsep}{4pt}
  \renewcommand{\arraystretch}{1.18}
  \caption{\textbf{Projects in DeliverBench.} Identifiers are used consistently throughout the paper; Appendix~\ref{app:project-profiles} provides detailed profiles.}
  \label{tab:project-names}
  \begin{tabularx}{\linewidth}{@{}lX@{\hspace{2em}}lX@{}}
    \toprule
    \textbf{ID} & \textbf{Project} & \textbf{ID} & \textbf{Project} \\
    \midrule
\textbf{P01} & Longtoutou Strategy & \textbf{P07} & Qixiang Financial Due Diligence \\
\textbf{P02} & Project Mars Refinancing & \textbf{P08} & Pharmacovigilance ICSR Processing \\
\textbf{P03} & Starbridge ITGC Audit & \textbf{P09} & Pharmaceutical Commercial Compliance \\
\textbf{P04} & Residential Interior Design & \textbf{P10} & StarPulse Legal Documentation \\
\textbf{P05} & Project Boomerang Forensics & \textbf{P11} & Evaluation Data Delivery \\
\textbf{P06} & IF Quantitative Research &  &  \\
    \bottomrule
  \end{tabularx}
\end{table}

\subsection{Project Sourcing and Benchmark Construction}
\label{sec:construction}

\begin{figure}[htbp]
  \centering
  \includegraphics[width=\linewidth]{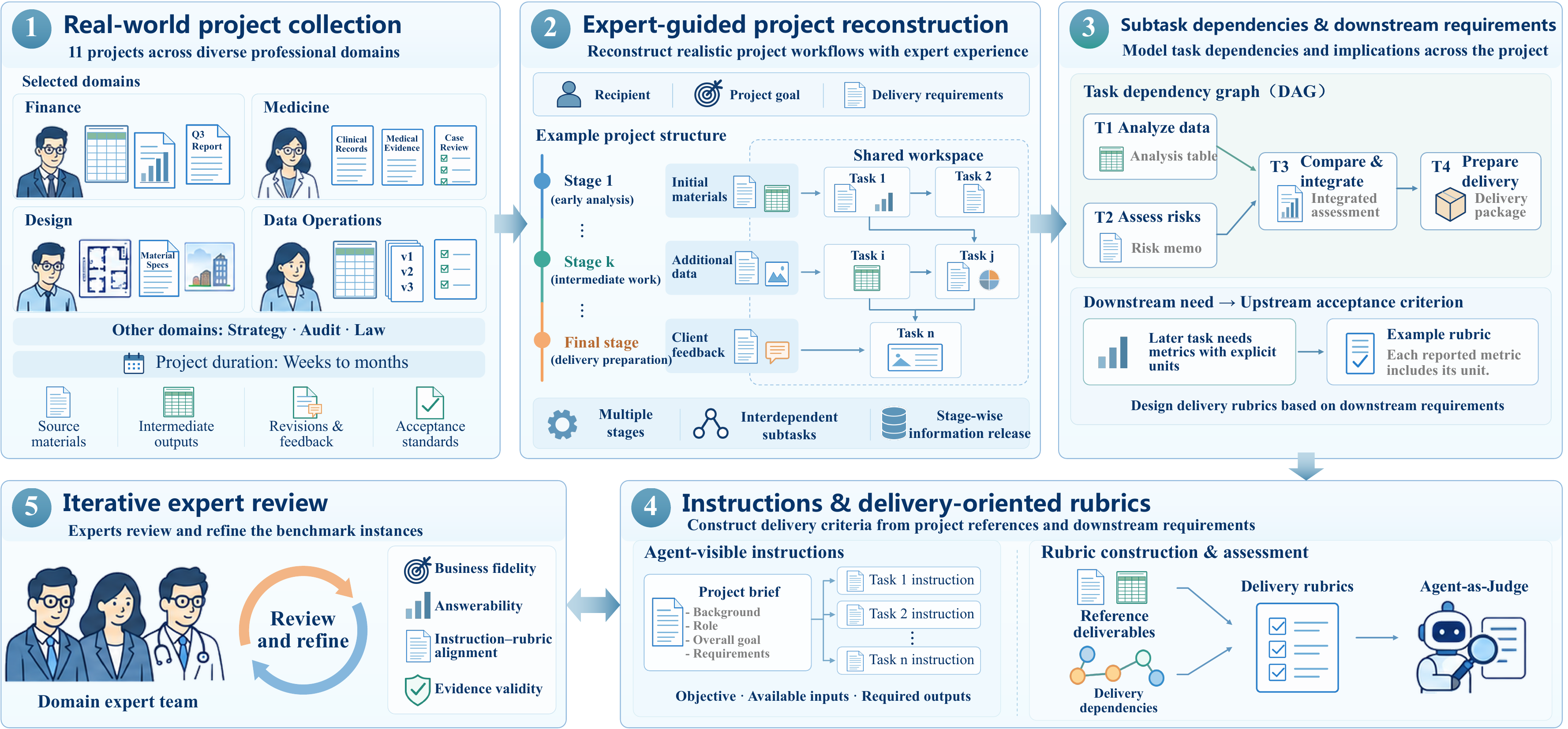}
  \caption{\textbf{Expert-guided construction of \benchmark{}.} Real-world materials are reconstructed into staged workflows. Artifact dependencies and downstream requirements guide instruction and rubric writing, followed by expert review of business fidelity, answerability, instruction--rubric alignment, and evidence validity.}
  \label{fig:construction}
\end{figure}

Figure~\ref{fig:construction} demonstrates the five-step process for reconstructing real-world projects as expert-reviewed benchmark instances.

\textbf{1. Collect real-world project materials.}
Domain experts (e.g., lawyers and investment bankers with several years of professional experience) provide projects spanning weeks to months, together with source records, intermediate and final deliverables, feedback, and acceptance standards.

\textbf{2. Reconstruct the project workflow.}
Experts help translate project scope and milestones into stages and subtasks with specified inputs, outputs, and information releases.
The reconstruction specifies task instructions and the release schedule, while distinguishing original records, reconstructed materials, and scenario assumptions.
For interactive feedback, experts define business roles, available facts, and review requirements; the response mechanism uses the agent's actual requests and submissions to determine which information and comments to return.

\textbf{3. Model dependencies and downstream requirements.}
We identify how later subtasks use earlier artifacts and which outputs a revision affects.
These downstream needs determine what upstream deliverables must contain and inform their acceptance criteria.

\textbf{4. Write instructions and delivery rubrics.}
The project brief and task instructions specify objectives, available inputs, and required outputs.
Reference artifacts and dependency requirements inform subtask, cross-file, and package rubrics; evaluator-only references and unreleased materials are kept separate from currently visible inputs.

\textbf{5. Review and refine with domain experts.}
Experts check four properties: \emph{business fidelity} to professional goals and constraints; \emph{answerability} from the materials visible to the current subtask; \emph{instruction--rubric alignment} between stated requirements and scoring; and \emph{evidence validity} for the facts and judgments used.
Unresolved evidence gaps require repair or exclusion.

\subsection{Execution and Evaluation Protocol}
\label{sec:protocol}

\begin{figure}[htbp]
  \centering
  \includegraphics[width=\linewidth]{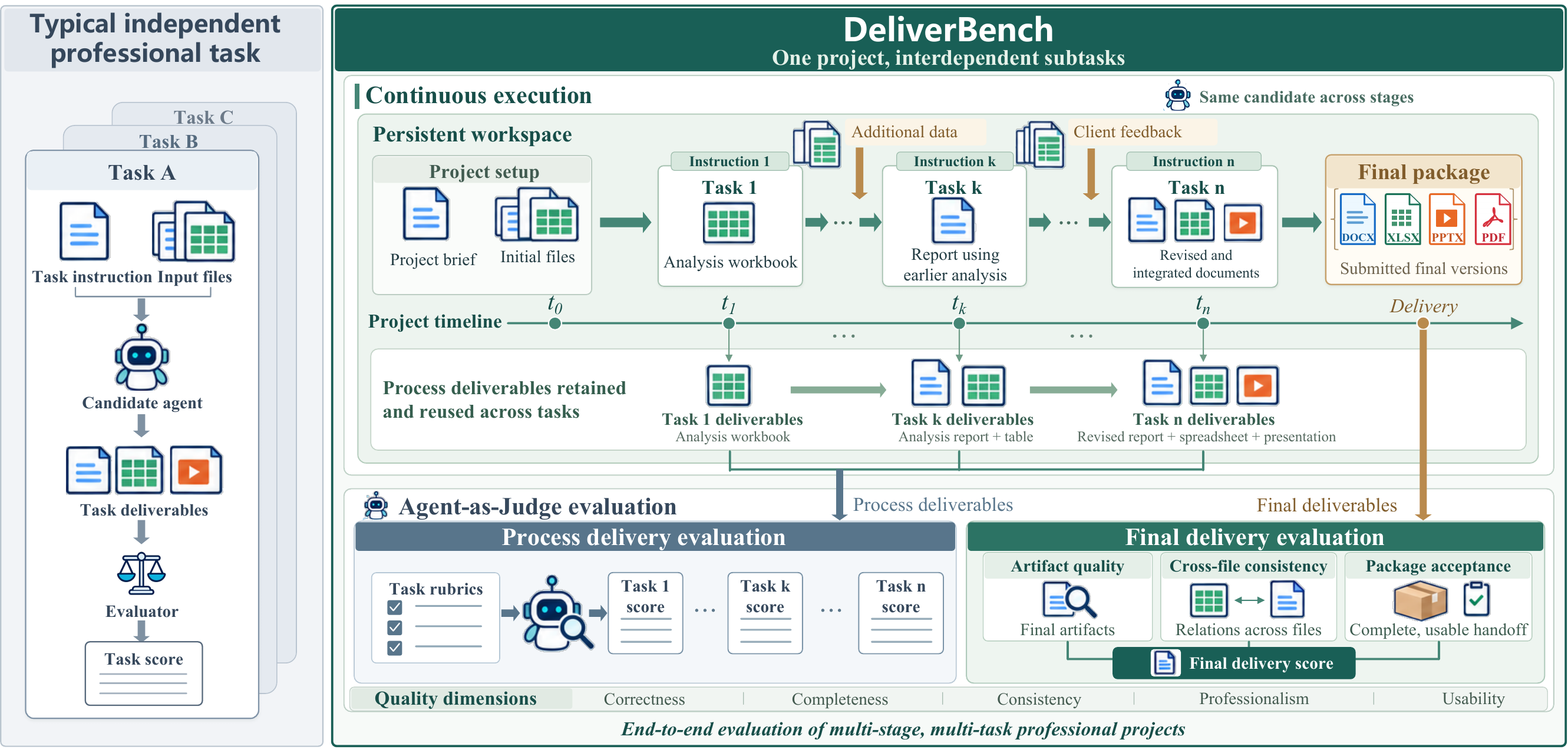}
  \caption{\textbf{Continuous project execution and delivery evaluation.} A typical task-level setup executes and scores assignments independently (left). DeliverBench keeps the same candidate in a persistent workspace as new data and feedback arrive; later subtasks use and revise its earlier artifacts (right). Process evaluation scores preserved subtask submissions. Final evaluation assesses artifact quality, cross-file consistency, and package acceptance on the submitted handoff. Quality-dimension labels provide a separate diagnostic view.}
  \label{fig:execution}
\end{figure}

Figure~\ref{fig:execution} contrasts continuous project execution with a typical task-level setup in which assignments are executed and scored independently.

\paragraph{Continuous execution and artifact inheritance.}
Execution starts from a project brief and initial files in a project-specific workspace.
The harness continues the agent's native session between subtasks and retains its working files; successive instructions extend the same project rather than initialize independent assignments.
Later subtasks use the agent's earlier deliverables as working inputs, without injected reference outputs.
Uncorrected errors therefore carry into downstream analyses, decisions, and the final package; the agent can also repair upstream artifacts while completing the work that depends on them.
Each submitted version is preserved for evaluation, while working files can be revised for subsequent use.
Agent environments are isolated across system--project runs, and private evaluator materials remain outside the agent-facing workspace.

\paragraph{New materials and feedback.}
Instructions and additional materials are released at designated points; unreleased sources remain unavailable.
At interactive events, simulated clients and reviewers respond to the agent's actual requests and submitted artifacts within role-specific facts and review requirements.
The schedule determines when an interaction occurs, while the submitted work determines which information or review comments are returned.
The agent must identify the calculations, conclusions, decisions, and handoff actions affected by new evidence and update their current versions.
Appendix~\ref{app:representative-workflow} illustrates this mechanism through lender feedback on the actual circulated documents in P02.

\paragraph{Process and final-delivery evaluation.}
Process evaluation assesses what the agent submitted after each subtask, using the information available at that point.
Final evaluation assesses the versions handed over at project completion and whether they jointly satisfy the delivery requirements.
Preserving intermediate submissions records the quality of work at each handoff, while final evaluation captures the package after subsequent work and repair.
Final evaluation has three targets: \emph{artifact quality} checks individual final files; \emph{cross-file consistency} checks agreement between related files; and \emph{package acceptance} checks the complete collection against the project objective, required deliverables, and handoff conditions.
An outdated value in an operative calculation, conflicting conclusions in related documents, and an unclear next action for the recipient illustrate these distinct scopes.
For project $p$ with $T_p$ subtasks, subtask artifacts $\mathcal{D}_{p,t}$, and final package $D_p$, we report
\begin{equation}
  \mathrm{Process}_p = \frac{1}{T_p}\sum_{t=1}^{T_p}\mathcal{S}(\mathcal{D}_{p,t},\mathcal{R}_{p,t}),
  \qquad
  \mathrm{Final}_p = \mathcal{S}(D_p,\mathcal{R}^{\mathrm{final}}_p),
  \label{eq:delivery-scores}
\end{equation}
where $\mathcal{R}_{p,t}$ contains subtask criteria and $\mathcal{R}^{\mathrm{final}}_p$ combines criteria for the three final-delivery targets.

\paragraph{Rubric-based judgment and scoring.}
Each criterion specifies a requirement, target artifacts, judgment evidence, and weight.
The five quality dimensions---correctness, completeness, consistency, professionalism, and usability---describe the property checked within these scopes, rather than additional evaluation targets.
We use an Agent-as-Judge approach~\citep{zhuge2024agentasjudge}: the judge assesses criteria from instructions and artifact evidence, supported by extraction and integrity checks; programmatic scoring applies the declared weights, dependencies, and acceptance gates.
The scoring operator $\mathcal{S}$ applies the scoring rules associated with each rubric set.
For final delivery, all applicable criteria enter a single aggregate: their signed integer-weighted judgments are summed, divided by the total available positive weight, and scaled to $[0,100]$, with out-of-range values clipped.
A triggered project-level fatal gate sets the final score to zero.
Artifact quality, cross-file consistency, and package acceptance identify the assessment scope of each criterion. Their separately normalized scores support diagnosis; the final score is computed directly from individual criteria, without an additional layer of target weights.
Criterion weights are shared across systems within each project, and the benchmark score averages the 11 project scores equally (Appendix~\ref{app:score-aggregation}).
Final-artifact assessment reuses applicable subtask criteria; previous judgments are reused only when the relevant artifacts, criteria, and evaluation context are unchanged.

Appendix~\ref{app:evaluation} specifies evidence scopes, scoring, and rubric reuse; Appendix~\ref{app:judge-settings} gives evaluator settings, and Section~\ref{sec:evaluation-validation} reports evaluator validation.

\subsection{Dataset Composition and Task Characteristics}
\label{sec:dataset-quality}

\begin{figure}[htbp]
  \centering
  \includegraphics[width=\linewidth]{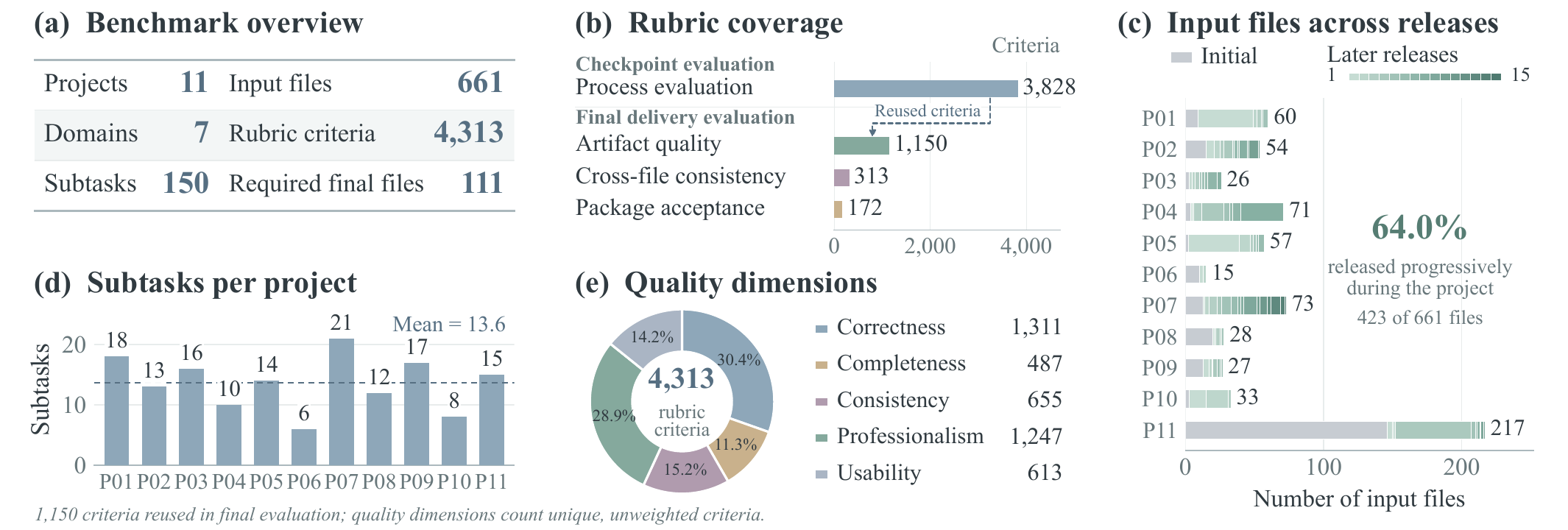}
  \caption{\textbf{Dataset overview.} (a) Benchmark scale, excluding duplicate input copies; deliverables are required final files. (b) Rubric coverage: 1,150 process criteria are reapplied to final deliverables. (c) Input releases, indexed within each project. (d) Subtask counts and their mean. (e) Unique criteria by primary quality dimension, without weighting. P01--P11 identify the same projects across panels.}
  \label{fig:dataset-overview}
\end{figure}

Figure~\ref{fig:dataset-overview} summarizes 11 projects, 150 subtasks, 661 input files, and 111 required final deliverables across seven domains: Finance, Medicine, Design, Data Operations, Strategy, Audit, and Law. Finance contains four projects, Medicine two, and each remaining domain one. Appendix~\ref{app:project-profiles} describes their workflows and deliverables.

The materials place different demands on agents. Finance combines transaction documents, financial workbooks, and market data, requiring numerical reconciliation and reproducible analysis. Medicine links clinical evidence and follow-up to safety reporting, and departmental feedback to compliance policies. Law requires agreements to reflect changing business facts; Strategy and Audit connect recommendations and control judgments to supporting evidence. Design requires visual materials and changing client requirements to be reconciled, while Data Operations emphasizes programmatic verification and version tracking.

Projects contain 6--21 subtasks. Overall, 423 of the 661 inputs (64.0\%) arrive after initialization, with 2--15 subsequent releases per project. Agents must therefore revise accumulated work as the information changes and carry the revisions into dependent deliverables. Appendix~\ref{app:project-profiles} gives project-specific materials, workflows, and output requirements.

The rubric inventory contains 4,313 unique criteria: 3,828 process criteria, 313 cross-file consistency criteria, and 172 package acceptance criteria. Final artifact-quality evaluation reapplies 1,150 process criteria to final versions, counted once in this inventory. The quality-dimension distribution describes annotation coverage rather than scoring weights.

\Needspace{7\baselineskip}
\section{Experiments and Analysis}
\label{sec:experiments}

\subsection{Experimental Setup and Metrics}
\label{sec:experimental-setup}
\textbf{Agent systems and execution.}
We evaluate seven model--harness systems: GPT-5.6 Sol with Codex, Claude Opus 5 with Claude Code, Gemini 3.8 Flash with Gemini CLI, Qwen 3.8 Max with Qwen Code, DeepSeek V4.1 Flash and GLM 5.2 with Claude Code, and Kimi K3 with Kimi CLI.
Projects follow the continuous protocol in Section~\ref{sec:protocol}, using Harbor~0.20.0 and Docker with a four-hour execution limit per subtask. System versions and execution settings are specified in Appendix~\ref{app:agent-systems}.

\textbf{Evaluation and reporting.}
GPT-5.6 Luna, running through Codex CLI~0.142.4, judges rubric criteria; deterministic scoring produces separate process and final-delivery scores (Equation~\ref{eq:delivery-scores}). We evaluate all 11 projects, yielding 77 system--project outcomes and 150 subtask submissions per system.
Process and final-delivery scores are averaged equally across projects; the process score first averages subtasks within each project.
We additionally report the fraction of subtask submissions scoring at least 90, pooling all 150 subtasks per system with equal weight.
This thresholded subtask pass rate measures the frequency of high-scoring intermediate submissions, rather than final-package acceptance.
For final delivery, Figure~\ref{fig:main-leaderboard} also reports the fraction of the 11 projects scoring at least 75.
Each system--project pair contributes one reported outcome; Appendix~\ref{app:run-provenance} specifies failure and recovery handling.
Appendix~\ref{app:evaluation} specifies scoring, rubric reuse, project means, and subtask pass rates.

\subsection{Main Results and Analysis}
\label{sec:main-results}
\label{sec:local-global}
\label{sec:analysis}
\begingroup
\setlength{\intextsep}{5pt}
\setlength{\textfloatsep}{8pt}
Figure~\ref{fig:main-leaderboard} summarizes overall performance across seven agent systems and 11 projects. Figure~\ref{fig:process-delivery} compares process and final-delivery scores at the system and project levels, while Figure~\ref{fig:delivery-diagnostics} presents quality profiles and resource use. Appendix~\ref{app:project-scores} provides the complete final-delivery and process-score matrices.

\textbf{Current agents struggle with reliable end-to-end project delivery.}
Final-delivery scores range from 40.4 to 65.0 out of 100, with Claude Opus 5 scoring 65.0 and GPT-5.6 Sol scoring 63.7 (Figure~\ref{fig:main-leaderboard}).
At the 90-point threshold, subtask pass rates range from 13.3\% to 50.0\%: Claude passes 75 of 150 subtasks, GPT passes 54 (36.0\%), and each remaining system passes at most 26.7\%.
Claude therefore clears the threshold on 21 more subtasks than GPT, a 14.0-percentage-point difference, yet still falls short on half of its submissions.
These results expose two delivery demands: producing reliable intermediate work and turning that work into a coherent final package.

\begin{figure}[!htbp]
  \centering
  \captionsetup{skip=5pt}
  \includegraphics[width=\linewidth]{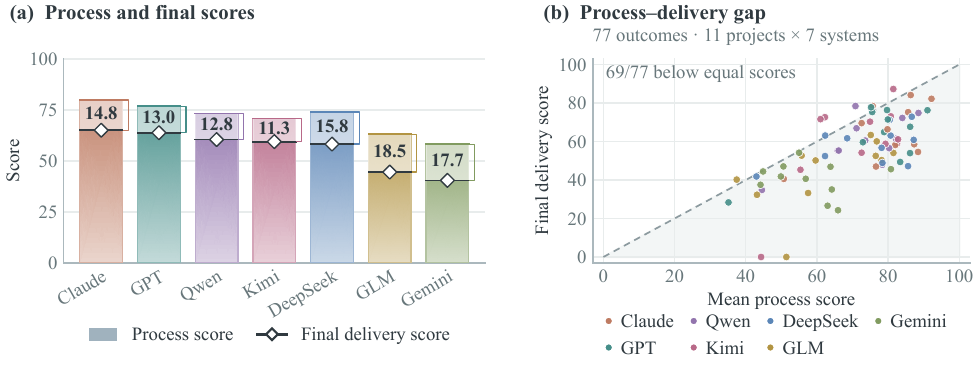}
  \caption{\textbf{Process and final-delivery scores.} (a) Mean process and final-delivery scores for each system; numbers inside the highlighted intervals give their differences in points. (b) All 77 system--project outcomes, with mean subtask score on the horizontal axis and final-delivery score on the vertical axis; the dashed line denotes equal scores. System-level means give equal weight to projects. Final-delivery scores retain fatal gates.}
  \label{fig:process-delivery}
\end{figure}

\textbf{Local task scores do not establish final-package readiness.}
Every system scores higher on process evaluation than on final delivery, with delivery gaps of 11.3--18.5 points (Figure~\ref{fig:process-delivery}(a)).
For example, Claude scores 79.8 on intermediate work and 65.0 on final delivery, while GLM scores 63.0 and 44.5, respectively.
This pattern also holds beyond system averages: 69 of 77 system--project outcomes lie below the equal-score line in Figure~\ref{fig:process-delivery}(b).
On Residential Interior Design (P04), Gemini averages 80.8 across subtask submissions but scores 45.7 on final delivery, a 35.1-point gap; on Qixiang Financial Due Diligence (P07), Claude averages 88.4 but delivers a final package scoring 54.6, a 33.8-point gap.
System-level rankings broadly agree, but the project-level view reveals requirements that remain unmet at handoff even after high-scoring intermediate work.
Averaging individually scored submissions does not establish whether the final versions agree across files or jointly meet handoff requirements.
Project-level evaluation checks these relationships and package conditions alongside the final artifacts themselves.
Section~\ref{sec:case-failures} identifies the concrete revision and handoff failures behind this distinction.

\FloatBarrier
\begin{figure}[!htbp]
  \centering
  \captionsetup{skip=5pt}
  \includegraphics[width=\linewidth]{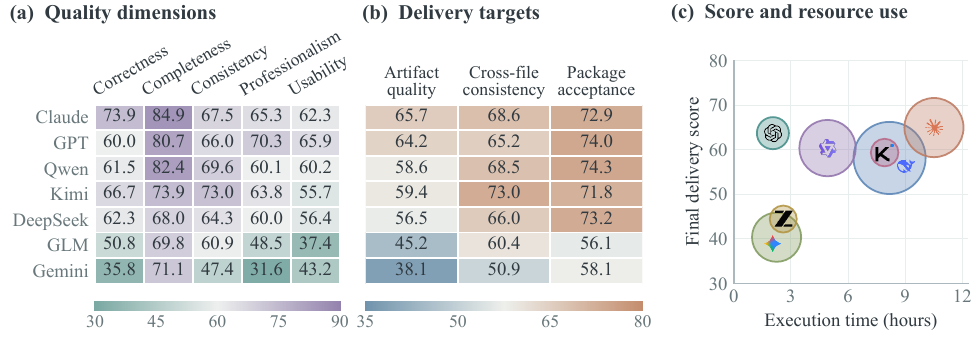}
  \caption{\textbf{Final-delivery quality and resource use.} (a) Five quality dimensions (color scale: 30--90). (b) Three final-delivery targets: artifact quality, cross-file consistency, and package acceptance (color scale: 35--80). (c) Mean project execution time versus final-delivery score; bubble area encodes mean total tokens, and logos identify the systems. The two heatmaps share the model order. System-level means give equal weight to projects. The diagnostic profiles in (a,b) precede project-level fatal gates; the final-delivery scores in (c) retain them.}
  \label{fig:delivery-diagnostics}
\end{figure}

\textbf{Content coverage exceeds operational quality.}
Completeness is the highest-scoring quality dimension for every system (68.0--84.9; mean 75.8), while usability averages 54.4, professionalism 57.1, correctness 58.7, and consistency 64.1 (Figure~\ref{fig:delivery-diagnostics}(a)).
The mean gap between completeness and usability is 21.4 points: covering requested content does not ensure that the recipient can use it for the intended work.
These five quality dimensions cut across the three final-delivery targets.

\textbf{Delivery requirements need separate diagnostic views.}
Figure~\ref{fig:delivery-diagnostics}(b) reports artifact quality, cross-file consistency, and package acceptance.
Their system-level means range from 38.1--65.7, 50.9--73.0, and 56.1--74.3, respectively.
Cross-file agreement alone cannot establish correctness: several files may consistently reuse the same unsupported upstream value. Artifact quality checks its evidential basis, cross-file consistency checks agreement across the resulting documents, and package acceptance checks whether the combined deliverables meet the recipient's requirements.
These averages do not locate the failure within an individual project.
For example, Claude's Project Boomerang Forensics (P05) package scores 88.8 on artifact quality and 100.0 on cross-file consistency, but only 60.0 on package acceptance: it leaves authorization contingencies, evidence-closure conditions, and the review-to-issuance sequence insufficiently specified for the recipient.
In Starbridge ITGC Audit (P03), all seven systems satisfy the four criteria directly addressing the incoming review issues, yet only four satisfy the associated propagation check and only one satisfies the associated cross-file traceability check.
Recording feedback, updating all affected artifacts, and handing over an actionable package are therefore distinct demands.

\Needspace{6\baselineskip}
\textbf{Greater resource use does not consistently yield better delivery.}
Following joint evaluation of agent performance and cost~\citep{kapoor2025agents}, Figure~\ref{fig:delivery-diagnostics}(c) reports mean project time of 2.05--10.51 hours and token usage of 26.0M--184.5M.
GPT and Claude score 63.7 and 65.0, respectively, but use 2.05 versus 10.51 hours and 36.4M versus 124.1M tokens per project. Claude uses 5.1 times as much execution time and 3.4 times as many tokens for a 1.3-point higher final score. Appendix~\ref{app:resource-use} reports the exact system means and project-level values under the accounting rules in Appendix~\ref{app:resource-accounting}.

\FloatBarrier
\endgroup
\suppressfloats[t]
\subsection{Case Studies}
\label{sec:case-studies}
We examine preserved artifacts to identify concrete delivery failures, then use subtask-score trajectories to study quality stability and recovery.

\subsubsection{Revision and Handoff Failures}
\label{sec:case-failures}
\begin{figure}[!htbp]
  \centering
  \includegraphics[width=\linewidth]{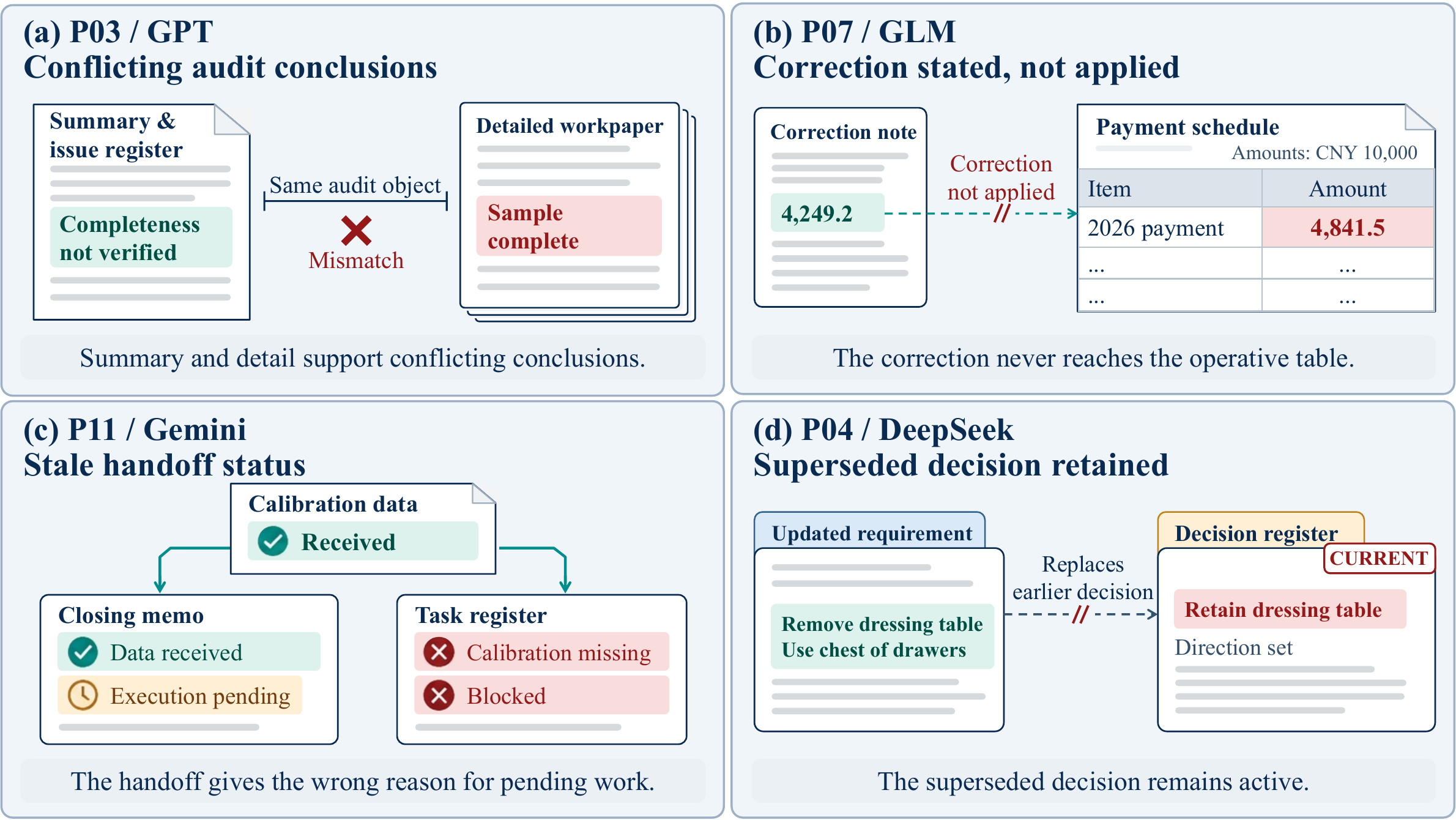}
  \caption{\textbf{Representative failures in maintaining a coherent delivery package.} (a) P03/GPT: the audit summary and issue register acknowledge an unverified population, while the detailed workpaper retains a positive completeness conclusion. (b) P07/GLM: a correction note records the revised payment amount, but the operative schedule retains the old amount (units: CNY 10,000). (c) P11/Gemini: the closing memo acknowledges receipt of calibration data, while the task register still treats the data as missing. (d) P04/DeepSeek: an updated requirement replaces a dressing table, but its retention remains an active decision. The figure summarizes inspected artifact evidence.}
  \label{fig:delivery-failures}
\end{figure}

Figure~\ref{fig:delivery-failures} presents four representative cases in which information was available or explicitly recorded, but its implications did not reach the relevant artifacts or current project state.

\textbf{Conflicting conclusions in Starbridge ITGC Audit (P03/GPT).}
After review, the audit summary and issue register state that population completeness and period coverage are unverified, preventing an operating-effectiveness conclusion.
The detailed workpaper for the same control still claims a complete sample and an appropriate random selection.
The revised limitation has reached the summary, but the supporting detail continues to assert an incompatible evidence state.
A recipient cannot obtain a consistent audit conclusion by following the package from its summary to its supporting workpaper.

\textbf{An unapplied correction in Qixiang Financial Due Diligence (P07/GLM).}
New feedback corrects a 2026 payment amount from 4,841.5 to 4,249.2 in units of CNY 10,000.
The revised workbook records the correct amount in its change note, but retains 4,841.5 in the operative capital-expenditure schedule, where it continues to feed the calculation.
Recognizing a correction is insufficient when the table used for subsequent financial analysis still carries the superseded value.

\textbf{Stale status in Evaluation Data Delivery (P11/Gemini).}
The closing memo records that calibration data have arrived and that the remaining model execution belongs to the next scheduled window.
The task register nevertheless lists missing calibration data as the current blocker.
This inconsistency changes the apparent next action: the recipient is told to obtain an input that is already available instead of completing the outstanding execution.

\textbf{A superseded decision in Residential Interior Design (P04/DeepSeek).}
The agent records a new requirement to remove the dressing table and use a chest of drawers.
Its final decision register still treats retention of the dressing table as the current direction and preserves related follow-up actions.
The failure is in maintaining the active decision state: entering the new requirement has not retired the decision that it replaces.

Together, these cases show that carrying work forward requires preserving the implications of new information across calculations, conclusions, decisions, and handoff actions.
A correction must reach the operative calculation; an evidence limitation must restrict the conclusions drawn from it; and a revised decision must change the instructions passed to the recipient.
Recording the update without changing these dependent artifacts leaves downstream work on the earlier basis.
The final package must communicate one coherent state that the recipient can act on.

\subsubsection{Delivery Stability and Artifact Recovery}
\label{sec:case-trajectories}
\begin{figure}[htbp]
  \centering
  \includegraphics[width=\linewidth]{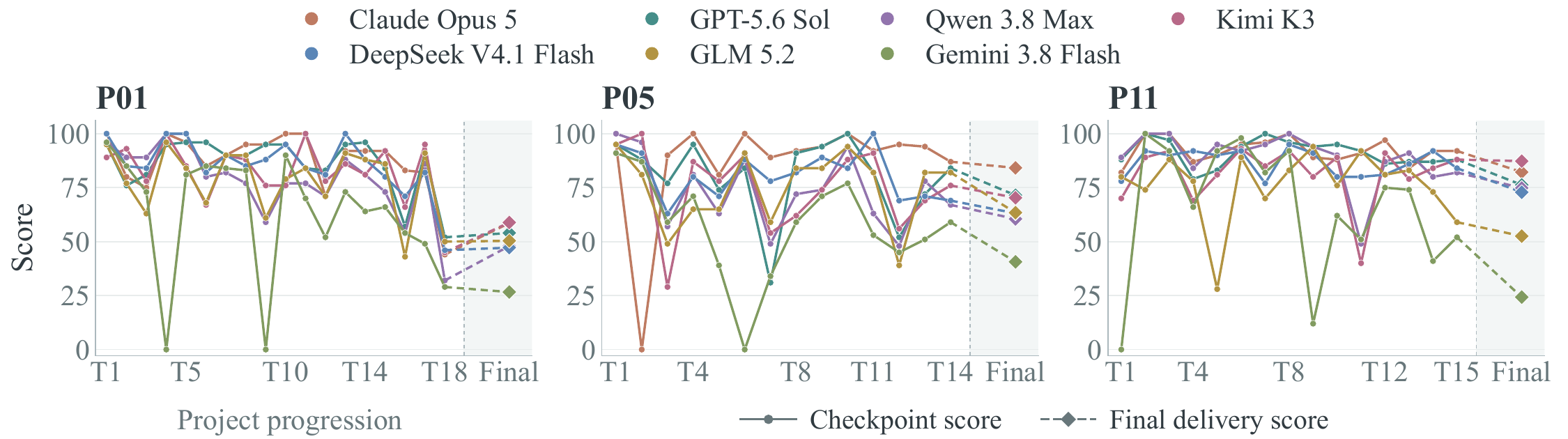}
  \caption{\textbf{Agent trajectories on three selected projects.} Circles show subtask scores; diamonds show final-delivery scores. Dashed terminal segments mark the change in evaluation scope. All subtask submissions are retained on a common score scale. The cases illustrate quality changes and recovery behaviors during project delivery.}
  \label{fig:case-trajectories}
\end{figure}

We analyze agent execution trajectories from Longtoutou Strategy (P01), Project Boomerang Forensics (P05), and Evaluation Data Delivery (P11).
Figure~\ref{fig:case-trajectories} pairs subtask scores with final-delivery scores for all seven agents.
Together with artifact and execution-trace inspection, this supports fine-grained analysis of how artifact quality changes during a project, where failures arise, and whether later work repairs earlier artifacts.
Appendix~\ref{app:project-trajectories} provides the corresponding trajectories for all 11 projects.

\textbf{Final delivery builds on intermediate artifacts and introduces distinct quality requirements.}
The final package inherits earlier deliverables, which agents must maintain, revise, and integrate as the project unfolds.
Its score is not a simple average of subtask scores: final evaluation examines the resulting artifact versions, cross-file relationships, and package acceptance.
The trajectories reveal recovery after an early failure in P05 and contrasting stability patterns in P11.
Project-delivery evaluation thus examines whether agents maintain accumulated work and transform it into a coherent, usable final package.

\textbf{Stronger agents generally sustain more stable delivery quality.}
For example, the Claude agent maintains scores of 82--100 throughout P11, whereas the Gemini agent ranges from 0 to 100 and suffers another sharp drop at T09.
The Gemini agent also exhibits substantial fluctuations in P01 and P05, and obtains the lowest final-delivery score in all three cases: 26.6, 40.6, and 24.3, respectively.
Its occasional high-scoring subtasks therefore do not translate into sustained quality across the project or a high-scoring final package.

\Needspace{5\baselineskip}
\textbf{Beyond stability, stronger agents exhibit a second core delivery capability: repairing intermediate artifacts.}
In P05, the Claude agent scores 0 at the second subtask because the required transaction-reconciliation workbook was not submitted.
During the third subtask, it detects the missing earlier deliverable and rebuilds the workbook from retained parsing results so that downstream entity-analysis references can resolve.
The rebuilt workbook contains 3,327 transaction rows and is preserved in the third-subtask archive.
The agent subsequently obtains the highest final-delivery score on P05, 84.1, compared with 40.6 for the Gemini agent.
By repairing the upstream workbook, the agent restores the input needed for downstream analysis and makes its transaction records available through the references used by later work. The execution trace thus shows recovery of a dependency: subsequent analysis can continue from a usable upstream artifact.

\subsection{Evaluation Reliability}
\label{sec:evaluation-validation}
Comparisons with human judgments provide an empirical check on model-based evaluation~\citep{zheng2023judging}.
We assess evaluator reliability by comparing expert and evaluator judgments on final deliverables produced by GPT-5.6 Sol--Codex for three projects: P01, P05, and P10.
Across 478 rubric decisions, the judgments agree on 446 (93.31\%); agreement is 93.04\% for artifact quality, 95.00\% for cross-file consistency, and 92.45\% for package acceptance.
Agreement is the fraction of matching binary judgments, pooled across criteria without delivery-score weights.
This validation covers final-delivery judgments for one agent system, including source-task criteria applied to its final artifacts.

\section{Conclusion and Limitations}
\label{sec:conclusion}

\benchmark{} evaluates agents on multi-stage professional projects reconstructed with domain experts. Later subtasks directly reuse the agent's earlier deliverables, making the quality of upstream work consequential for downstream analysis and final handoff. It assesses intermediate submissions and final packages as agents incorporate new information and revise earlier artifacts in a persistent workspace.

Across seven model--harness systems, the highest mean final-delivery score is approximately 65 out of 100, and the highest subtask pass rate is 50\% at the 90-point threshold. Every system has a lower mean final-delivery score than mean process score, and 69 of 77 system--project outcomes exhibit this gap. Case studies reveal conflicting conclusions, unapplied corrections, stale handoff states, and superseded decisions; they also show the value of repairing missing upstream artifacts. Reliable delivery requires tracing the effects of new evidence, translating analytical findings into current decisions, and repairing upstream artifacts together with the downstream work that uses them. Evaluating final packages alongside intermediate submissions makes these relationships and acceptance requirements explicit. More reliable project delivery could broaden enterprise delegation and support one-person companies, where an individual sets business goals and oversees agents through final handoff.

\label{sec:limitations}
Our study has several limitations. Coverage remains limited to eleven projects across seven domains. Scheduled input releases and role-specific response rules support reproducible evaluation but do not fully capture open-ended client communication or live business decisions. Future work can broaden project coverage, introduce richer interaction, and extend expert validation to assess delivery capability more comprehensively in professional settings.

\clearpage

\begingroup
\small
\setlength{\bibsep}{0.5pt}
\bibliography{references}

\begin{thebibliography}{21}
\providecommand{\natexlab}[1]{#1}
\providecommand{\url}[1]{\texttt{#1}}
\expandafter\ifx\csname urlstyle\endcsname\relax
  \providecommand{\doi}[1]{doi: #1}\else
  \providecommand{\doi}{doi: \begingroup \urlstyle{rm}\Url}\fi

\bibitem[Boisvert et~al.(2024)Boisvert, Thakkar, Gasse, Caccia, De~Chezelles,
  Cappart, Chapados, Lacoste, and Drouin]{boisvert2024workarenapp}
L{\'e}o Boisvert, Megh Thakkar, Maxime Gasse, Massimo Caccia, Thibault~L
  De~Chezelles, Quentin Cappart, Nicolas Chapados, Alexandre Lacoste, and
  Alexandre Drouin.
\newblock Workarena++: Towards compositional planning and reasoning-based
  common knowledge work tasks.
\newblock \emph{Advances in Neural Information Processing Systems},
  37:\penalty0 5996--6051, 2024.

\bibitem[Finley et~al.(2026)Finley, Panavas, Mikkelson, Hinton, Ganss, Monton,
  Kendall, Spradlin, Bye, O'Brien, Ylvisaker, Ray, Garre, Mehta, and
  Chen]{finley2026dayjob}
Stephanie Finley, Liudas Panavas, Thomas Mikkelson, Cam Hinton, Stacey Ganss,
  Bradley Monton, Emily Kendall, Michelle Spradlin, Lydia Bye, Michael O'Brien,
  Lauren Ylvisaker, Derek Ray, Suhaas Garre, Sushant Mehta, and Edwin Chen.
\newblock {DAYJOB}: A benchmark for long-horizon professional work.
\newblock \emph{arXiv preprint arXiv:2610.01306}, 2026.
\newblock \doi{10.48550/arXiv.2610.01306}.
\newblock URL \url{https://arxiv.org/abs/2610.01306}.

\bibitem[Kapoor et~al.(2024)Kapoor, Stroebl, Siegel, Nadgir, and
  Narayanan]{kapoor2025agents}
Sayash Kapoor, Benedikt Stroebl, Zachary~S Siegel, Nitya Nadgir, and Arvind
  Narayanan.
\newblock Ai agents that matter.
\newblock \emph{arXiv preprint arXiv:2407.01502}, 2024.

\bibitem[Li et~al.(2026)Li, Feng, Xu, Ma, Zheng, Jiang, Sun, Shao, Chen, Huang,
  Han, Lee, Xu, Zeng, Hua, Zhang, Alomair, Krishna, Zettlemoyer, Koh,
  Ramasubramanian, Niu, Yue, and Poovendran]{li2026jobbench}
Yuetai Li, Yichen Feng, Zhangchen Xu, Zixian Ma, Kaiyuan Zheng, Fengqing Jiang,
  Xinghua Sun, Rulin Shao, Zichen Chen, Yue Huang, Xinyang Han, Brian Lee,
  Kayla Xu, Shenglai Zeng, Hang Hua, Xiangliang Zhang, Basel Alomair, Ranjay
  Krishna, Luke Zettlemoyer, Pang~Wei Koh, Bhaskar Ramasubramanian, Luyao Niu,
  Xiang Yue, and Radha Poovendran.
\newblock {JobBench}: Aligning agent work with human will, 2026.
\newblock URL \url{https://arxiv.org/abs/2605.26329}.

\bibitem[Ma et~al.(2024)Ma, Zhang, Zhu, Yang, Yang, Jin, Lan, Kong, and
  He]{ma2024agentboard}
Chang Ma, Junlei Zhang, Zhihao Zhu, Cheng Yang, Yujiu Yang, Yaohui Jin,
  Zhenzhong Lan, Lingpeng Kong, and Junxian He.
\newblock Agentboard: An analytical evaluation board of multi-turn llm agents.
\newblock \emph{Advances in neural information processing systems},
  37:\penalty0 74325--74362, 2024.

\bibitem[Malone and Crowston(1994)]{malone1994coordination}
Thomas~W Malone and Kevin Crowston.
\newblock The interdisciplinary study of coordination.
\newblock \emph{ACM Computing Surveys (CSUR)}, 26\penalty0 (1):\penalty0
  87--119, 1994.

\bibitem[Patwardhan et~al.(2026)Patwardhan, Dias, Proehl, Kim, Wang, Watkins,
  Fishman, Aljubeh, Thacker, Fauconnet, et~al.]{patwardhan2025gdpval}
Tejal Patwardhan, Rachel Dias, Elizabeth Proehl, Grace Kim, Michele Wang,
  Olivia Watkins, Simon Fishman, Marwan Aljubeh, Phoebe Thacker, Laurance
  Fauconnet, et~al.
\newblock Gdpval: Evaluating ai model performance on real-world economically
  valuable tasks.
\newblock In \emph{International Conference on Learning Representations},
  volume 2026, pages 24005--24040, 2026.

\bibitem[Shepard and Salimans(2026)]{shepard2026automationbench}
Daniel Shepard and Robin Salimans.
\newblock {AutomationBench}.
\newblock \emph{arXiv preprint arXiv:2604.18934}, 2026.
\newblock URL \url{https://arxiv.org/abs/2604.18934}.

\bibitem[Starace et~al.(2025)Starace, Jaffe, Sherburn, Aung, Chan, Maksin,
  Dias, Mays, Kinsella, Thompson, et~al.]{starace2025paperbench}
Giulio Starace, Oliver Jaffe, Dane Sherburn, James Aung, Jun~Shern Chan, Leon
  Maksin, Rachel Dias, Evan Mays, Benjamin Kinsella, Wyatt Thompson, et~al.
\newblock Paperbench: evaluating ai's ability to replicate ai research.
\newblock \emph{arXiv preprint arXiv:2504.01848}, 2025.

\bibitem[Sun(2026)]{sun2026sqbenchbenchmarkevaluatingtask}
Summer Sun.
\newblock Sqbench: A benchmark for evaluating task delivery by language-model
  agents in production-oriented workflows.
\newblock \emph{arXiv preprint arXiv:2607.23123}, 2026.

\bibitem[Sun et~al.(2026)Sun, Han, Zhang, Pang, Wang, Cao, Huang, Duroiu,
  Zhang, Lin, et~al.]{sun2026ale}
Yiyou Sun, Xinyang Han, Weichen Zhang, Yuanbo Pang, Tianyu Wang, Yuhan Cao,
  Yixiao Huang, Chris Duroiu, Haoyun Zhang, Jeffrey Lin, et~al.
\newblock Agents' last exam.
\newblock \emph{arXiv preprint arXiv:2606.05405}, 2026.

\bibitem[Tang et~al.(2026)Tang, Zhou, Liu, Li, Wu, Wang, Huang, Zhou, Zhou,
  Song, et~al.]{tang2026workspacebench}
Zirui Tang, Xuanhe Zhou, Yumou Liu, Linchun Li, Yukai Wu, Weizheng Wang,
  Hongzhang Huang, Wei Zhou, Jun Zhou, Jiachen Song, et~al.
\newblock Workspace-bench 1.0: Benchmarking ai agents on workspace tasks with
  large-scale file dependencies.
\newblock \emph{arXiv preprint arXiv:2605.03596}, 2026.

\bibitem[Vidgen et~al.(2026)Vidgen, Mann, Fennelly, Stanly, Rothman, Burstein,
  Benchek, Ostrofsky, Ravichandran, Sur, et~al.]{vidgen2026apexagents}
Bertie Vidgen, Austin Mann, Abby Fennelly, John~Wright Stanly, Lucas Rothman,
  Marco Burstein, Julien Benchek, David Ostrofsky, Anirudh Ravichandran, Debnil
  Sur, et~al.
\newblock Apex-agents.
\newblock \emph{arXiv preprint arXiv:2601.14242}, 2026.

\bibitem[Wang et~al.(2024)Wang, Cui, Zhong, Zhang, Yin, Lin, and
  Shang]{wang2024officebench}
Zilong Wang, Yuedong Cui, Li~Zhong, Zimin Zhang, Da~Yin, Bill~Yuchen Lin, and
  Jingbo Shang.
\newblock Officebench: Benchmarking language agents across multiple
  applications for office automation.
\newblock \emph{arXiv preprint arXiv:2407.19056}, 2024.

\bibitem[Xu et~al.(2026)Xu, Song, Li, Tang, Jain, Bao, Wang, Zhou, Guo, Cao,
  et~al.]{xu2024theagentcompany}
Frank~Fangzheng Xu, Yufan Song, Boxuan Li, Yuxuan Tang, Kritanjali Jain,
  Mengxue Bao, Zora Wang, Xuhui Zhou, Zhitong Guo, Murong Cao, et~al.
\newblock Theagentcompany: benchmarking llm agents on consequential real world
  tasks.
\newblock \emph{Advances in Neural Information Processing Systems}, 38, 2026.

\bibitem[Yao et~al.(2024)Yao, Shinn, Razavi, and Narasimhan]{yao2024taubench}
Shunyu Yao, Noah Shinn, Pedram Razavi, and Karthik Narasimhan.
\newblock {$\tau$-bench}: A benchmark for tool-agent-user interaction in
  real-world domains.
\newblock \emph{arXiv preprint arXiv:2406.12045}, 2024.

\bibitem[Yuan et~al.(2026)Yuan, Zhou, Xiong, Wu, Sun, Song, Cui, Wang, Wu, Li,
  Lu, Lu, Zhen, Wang, Deng, Yang, Chen, Zheng, Su, Yu, Zou, Agashe, Lu, Kaur,
  Qi, Chen, Sala, Liu, Lin, Yu, Su, Reddy, Wang, Qi, Xie, and
  Yu]{yuan2026osworld2}
Mengqi Yuan, Zilong Zhou, Xinzhuang Xiong, Weiming Wu, Jiayang Sun, Jiamin
  Song, Kaiqian Cui, Bowen Wang, Haoyuan Wu, Yitong Li, Dunjie Lu, Haikong Lu,
  Qi~Zhen, Xinyuan Wang, Jiaqi Deng, Yuhao Yang, Cheng Chen, Boyuan Zheng, Alex
  Su, Xiao Yu, Hao Zou, Saaket Agashe, Xing~Han Lu, Manpreet Kaur, Zhengyang
  Qi, Vincent~Sunn Chen, Frederic Sala, Dayiheng Liu, Junyang Lin, Zhou Yu,
  Yu~Su, Siva Reddy, Xin~Eric Wang, Peng Qi, Tianbao Xie, and Tao Yu.
\newblock {OSWorld 2.0}: Benchmarking computer use agents on long-horizon
  real-world tasks.
\newblock \emph{arXiv preprint arXiv:2606.29537}, 2026.
\newblock \doi{10.48550/arXiv.2606.29537}.
\newblock URL \url{https://arxiv.org/abs/2606.29537}.

\bibitem[Zhang et~al.(2026)Zhang, Xu, Wang, Gui, Zhang, Lei, He, He, Qin, Zhu,
  et~al.]{zhang2026pibench}
Haoran Zhang, Luxin Xu, Zhilin Wang, Runquan Gui, Shunkai Zhang, Haodi Lei,
  Zihao He, Bingsu He, Chicheng Qin, Tong Zhu, et~al.
\newblock {$\pi$-Bench}: Evaluating proactive personal assistant agents in
  long-horizon workflows.
\newblock \emph{arXiv preprint arXiv:2605.14678}, 2026.

\bibitem[Zheng et~al.(2023)Zheng, Chiang, Sheng, Zhuang, Wu, Zhuang, Lin, Li,
  Li, Xing, et~al.]{zheng2023judging}
Lianmin Zheng, Wei-Lin Chiang, Ying Sheng, Siyuan Zhuang, Zhanghao Wu, Yonghao
  Zhuang, Zi~Lin, Zhuohan Li, Dacheng Li, Eric Xing, et~al.
\newblock Judging llm-as-a-judge with mt-bench and chatbot arena.
\newblock \emph{Advances in neural information processing systems},
  36:\penalty0 46595--46623, 2023.

\bibitem[Zhu et~al.(2026)Zhu, Ma, Liu, Wang, Zhang, Ding, Gu, Zhong, Meng, Gao,
  et~al.]{zhu2026startupbench}
Liya Zhu, Xin Ma, Tao Liu, Haodong Wang, Ge~Zhang, Jingzhe Ding, Qingshui Gu,
  Yongjie Zhong, Jinxiang Meng, Yuan Gao, et~al.
\newblock Startupbench: Benchmarking general-purpose agents on market-validated
  end-to-end workflows.
\newblock \emph{arXiv preprint arXiv:2608.17800}, 2026.

\bibitem[Zhuge et~al.(2024)Zhuge, Zhao, Ashley, Wang, Khizbullin, Xiong, Liu,
  Chang, Krishnamoorthi, Tian, et~al.]{zhuge2024agentasjudge}
Mingchen Zhuge, Changsheng Zhao, Dylan Ashley, Wenyi Wang, Dmitrii Khizbullin,
  Yunyang Xiong, Zechun Liu, Ernie Chang, Raghuraman Krishnamoorthi, Yuandong
  Tian, et~al.
\newblock Agent-as-a-judge: Evaluate agents with agents.
\newblock \emph{arXiv preprint arXiv:2410.10934}, 2024.

\end{thebibliography}
\bibliographystyle{plainnat}
\endgroup

\clearpage
\beginappendix
\section{Benchmark Comparison Details}
\label{app:comparison-definitions}
Table~\ref{tab:benchmark-comparison} summarizes support specified in published protocols: a checkmark denotes explicit support, a triangle limited or related support, and a cross support not specified. These marks describe benchmark design rather than model performance or the prevalence of a feature across instances.

\paragraph{Evaluation unit.}
The unit is the scope of one evaluated instance, rather than its surrounding environment, individual tool actions, or rubric criteria. A project world can contain independent professional tasks, as in APEX-Agents~\citep[Sections~3.1--3.3]{vidgen2026apexagents}. Multiple output files alone do not change a task into a staged project. The labels describe scope rather than impose a common duration or file count.

\paragraph{Evolving inputs/feedback.}
Full support requires new task-relevant materials, requirements, or substantive collaborator feedback during execution. Scheduled releases and simulated-manager feedback qualify; ordinary tool observations, errors, continuation prompts, and retrieval of pre-existing files do not. TheAgentCompany incorporates simulated-manager feedback~\citep[Section~3 and Appendix~C]{xu2024theagentcompany}; DeliverBench releases materials and client feedback across stages (Section~\ref{sec:protocol}).

\paragraph{Long-horizon tasks.}
Support follows an explicit benchmark objective or documented substantial compositional execution, without a common numerical step threshold. SQBench receives partial support because it mixes atomic capabilities with composite and domain workflows~\citep[Sections~3.2--3.4]{sun2026sqbenchbenchmarkevaluatingtask}.

\paragraph{Explicit subtasks and artifact handoff.}
Full subtask support requires instantiated component tasks within a workflow; steps, checkpoints, or hierarchical requirements within one assignment receive partial support. Full artifact-handoff support requires later, separately specified tasks to consume or revise the same agent's earlier outputs. Dependencies within one workflow, such as OfficeBench's spreadsheet-to-email sequence, receive partial support~\citep{wang2024officebench}. Dependencies among supplied inputs alone do not qualify. Workspace-Bench restores the workspace between independent tasks~\citep[Section~4.5]{tang2026workspacebench}.

\paragraph{Intermediate evaluation.}
Full support requires checks or scores for defined intermediate subtasks or milestones, whether evaluated online or retrospectively. WorkArena++ and TheAgentCompany provide such checks~\citep{boisvert2024workarenapp,xu2024theagentcompany}. Partial support covers partial-progress assessment without defined subtask or milestone checks, or general retrospective process checks. Retrospective checkpoint scoring does not itself establish separately preserved subtask submissions.

\paragraph{Project delivery.}
Full support requires evaluating an integrated multi-file submission against a shared project outcome, regardless of whether the benchmark calls the instance a task or project. DAYJOB's cross-deliverable consistency checks qualify~\citep[Section~3.2]{finley2026dayjob}, as does full PaperBench's repository reproduction protocol~\citep[Sections~2.1--2.2]{starace2025paperbench}. This column does not additionally require staged inputs. Task-level output checks receive partial support where a project-package acceptance protocol is unspecified; WorkArena++ and AutomationBench instead validate application task state.

\paragraph{OSWorld 2.0 and AutomationBench.}
The OSWorld~2.0 row follows its published 108-task protocol. Dynamic messages support evolving inputs, and task-specific workflow checkpoints support intermediate evaluation, although scored from final state. Its reimbursement case evaluates a claim together with three generated supporting documents against shared requirements, qualifying for project delivery. Within-workflow artifact reuse and dependent steps support partial handoff and subtask structure; they are not separately specified project tasks~\citep[Sections~2.1--2.2 and Appendix~H.1.1]{yuan2026osworld2}.
AutomationBench's dependent API operations, including creating an Asana record and subsequently updating it, support partial handoff and subtask structure, and its compositional business workflows support the long-horizon column. Its single initial instruction and pre-seeded records do not establish evolving inputs. Acceptance checks application state rather than a multi-file project package. Final-state assertion diagnostics support limited intermediate evaluation; official task success requires all assertions to pass~\citep[Sections~2--4 and 11]{shepard2026automationbench}.

\paragraph{Boundary cases and evidence.}
DAYJOB's post-attempt trajectory inspection and partial-credit reporting support partial intermediate evaluation, while its initial workspace and non-interactive attempt do not establish evolving inputs or separate artifact handoffs~\citep[Sections~3.1--3.2 and 4.1--4.4]{finley2026dayjob}. JobBench's chained criteria and multi-part outputs support partial subtask, intermediate, and package assessment~\citep[Sections~2.2--3.1 and Appendix~E]{li2026jobbench}. SQBench entries follow its published protocol because official tasks and the production scorer are withheld. The DeliverBench row follows Section~\ref{sec:benchmark}.

\FloatBarrier
\section{Project Example and Profiles}
\label{app:construction}
The eleven projects span Finance, Medicine, Design, Data Operations, Strategy, Audit, and Law. Each connects successive assignments through evolving materials, feedback, and agent-produced artifacts. The profiles use project identifiers P01--P11 and the project names in Table~\ref{tab:project-names}.

\subsection{Representative Project Workflow}
\label{app:representative-workflow}
\begin{figure}[!htbp]
  \centering
  \includegraphics[width=\linewidth]{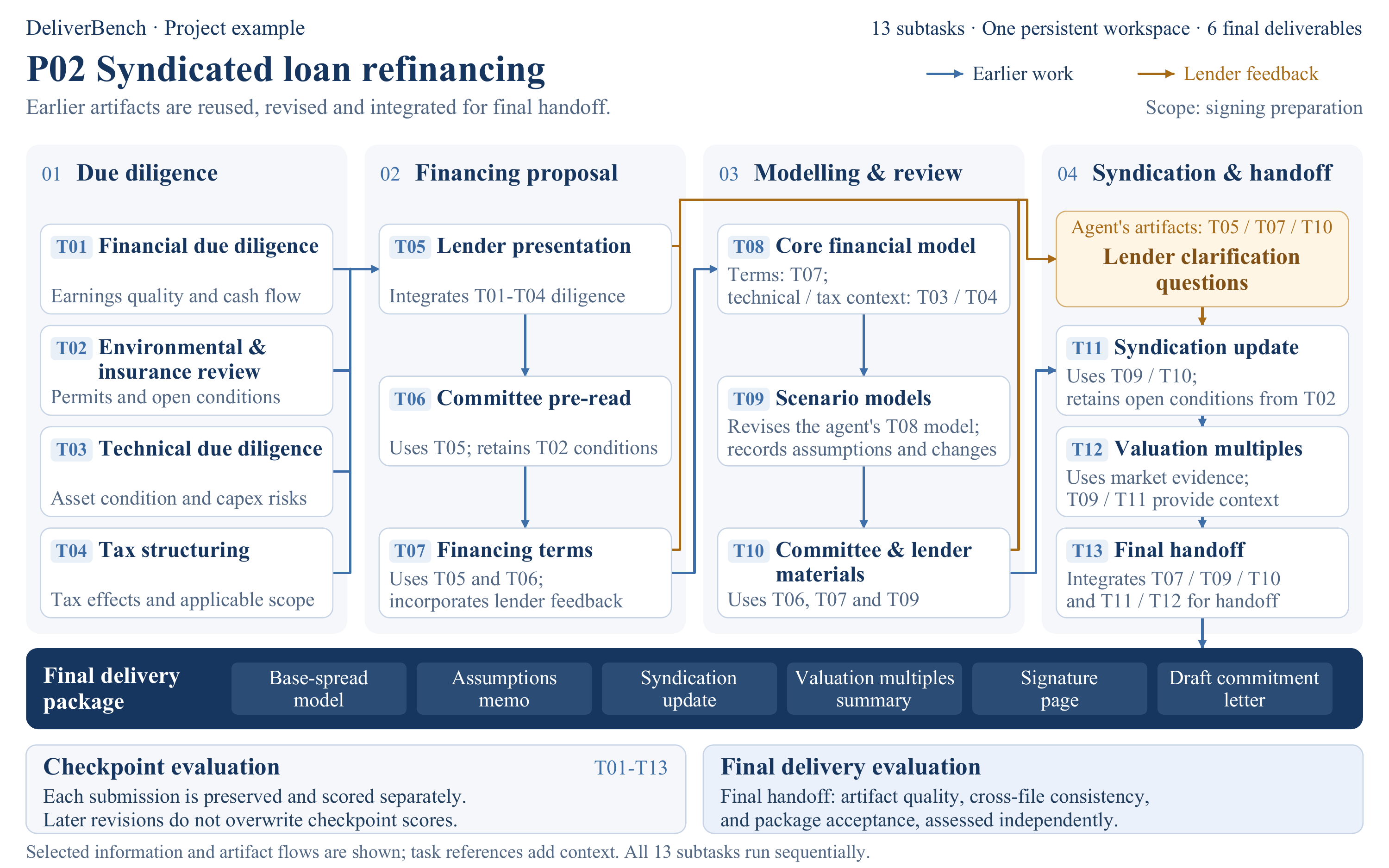}
  \caption{\textbf{Project Mars Refinancing (P02): directly dependent artifacts across four stages.} The thirteen sequential subtasks are organized into four stages. Blue arrows show selected flows of earlier work; the gold path shows lender clarification questions informed by the agent's submitted materials. The final package reuses and updates earlier models and reports alongside newly prepared signing documents.}
  \label{fig:project-02-workflow}
\end{figure}

Project Mars Refinancing (P02) follows a transaction team preparing syndicated refinancing for a digital infrastructure business. The agent turns due diligence into financing proposals, financial models, lender communications, and a six-file signing-preparation package. Figure~\ref{fig:project-02-workflow} groups its thirteen sequential subtasks into four stages. All subtasks share one persistent workspace: later subtasks use the agent's actual earlier files, including their assumptions, conclusions, and unresolved issues.

\paragraph{Stage 1: due diligence establishes the inputs (T01\textendash{}T04).}
T01 finalizes financial due diligence on earnings quality and cash available for debt service. T02 assesses environmental permits and insurance, identifying open conditions that affect financing readiness. T03 examines asset condition and capital-expenditure risks, and T04 updates the tax analysis for a separated business perimeter. These outputs establish the evidence, assumptions, and qualifications that the later financing work must use.

\paragraph{Stage 2: financing proposals build on the diligence (T05\textendash{}T07).}
T05 integrates T01--T04 into a lender presentation. T06 uses that presentation and the outstanding conditions to prepare an investment-committee pre-read. T07 then revises the financing terms using the earlier proposal, committee materials, a new rate notice, and lender feedback. A finding first recorded in due diligence must therefore remain reflected in the financing proposal and the conditions attached to it.

\paragraph{Stage 3: models and review materials carry the terms forward (T08\textendash{}T10).}
T08 updates the core financial model using T07's terms and the technical and tax context from T03--T04. T09 directly revises the agent's T08 workbook to produce the BaseSpread and CasePreferred scenarios, with an assumptions memorandum and a change log. T10 draws on the committee pre-read, financing terms, and scenario analysis to prepare the final committee pack, lender invitation, and working-group status note. This creates a concrete calculation dependency: a financing assumption left unchanged in the core model enters the scenario analysis, while the review and lender documents must reflect the resulting financing position.

\paragraph{Stage 4: lender coordination leads to final handoff (T11\textendash{}T13).}
Lender clarification questions are generated from the agent's actual T05 presentation, T07 terms, and T10 invitation. The reviewer checks the circulated documents against the permitted transaction sources and returns clarification points supported by their contents; the questions can therefore differ between agent submissions. T11 addresses those questions alongside new subscription and execution updates, drawing on the scenario analysis and retaining unresolved conditions from earlier diligence. T12 prepares valuation-multiple reference ranges using market evidence, with T09 and T11 providing transaction context. T13 assembles the signing-preparation package from this accumulated work and prepares the signature page and draft commitment letter.

\paragraph{The package preserves the dependency chain.}
The six final files are the BaseSpread model and assumptions memorandum from T09, the syndication update from T11, the valuation-multiples summary from T12, and the signature page and draft commitment letter from T13. Final copies may be revised to reflect subsequent information, with the basis for changes recorded. Their financing amounts, assumptions, open conditions, and commitment status must agree. An unresolved permit condition, for example, must remain visible from the diligence through lender coordination and signing preparation; an updated financing term must reach both the model that calculates its consequences and the documents that communicate it. P02 thus tests whether agents maintain these dependencies while producing a coherent final package.

\paragraph{What must survive each handoff.}
The dependency chain carries meaning and decision constraints as well as files. A debt figure used for refinancing must retain its borrower, facility scope, reporting date, and units; a consolidated borrowing total cannot simply become the balance of one facility. Financing scenarios must distinguish agreed terms from modelling assumptions, and that distinction must remain visible in the syndication update and signing drafts. If the model identifies a debt-service or covenant constraint, the subsequent recommendation must address its consequences for feasibility, conditions, or further work. Finally, each recipient needs a usable version of the package: outstanding approvals must be explicit, and internal working information must be separated from documents intended for external circulation.

\FloatBarrier

\Needspace{8\baselineskip}
\subsection{Project Profiles}
\label{app:project-profiles}

\noindent\textbf{P01. Longtoutou Strategy.} A strategy and capital advisory engagement supports a manufacturing company's operating-improvement and industrial-cooperation decisions. Financial baselines, site visits, partner profiles, and industry comparisons feed into alternative cooperation structures; new evidence and review comments require coordinated updates to analysis and recommendations. Delivery combines an analytical workbook, strategy and cooperation proposals, an investment memorandum, a management presentation, and handover notes identifying assumptions, evidence gaps, and decisions requiring approval.

\noindent\textbf{P02. Project Mars Refinancing.} A transaction team prepares syndicated refinancing for a digital infrastructure business. Financial, environmental, engineering, and tax due diligence inform financing terms, cash-flow models, and lender coordination. New evidence and lender responses must be reconciled across assumptions, financing amounts, valuation references, and commitment status. The signing-readiness package contains a scenario model, assumptions memorandum, syndication update, valuation presentation, and draft signing documents, distinguishing proposed terms, conditional commitments, and completed approvals.

\noindent\textbf{P03. Starbridge ITGC Audit.} An information-technology general-controls audit supports a financial audit team's reliance assessment. Planning, scoping, risk-control mapping, walkthroughs, and testing cover six control domains. Missing evidence and cross-domain review findings require coordinated revisions without presenting incomplete tests as effective controls. The package includes workpapers, an issue register, reliance conclusions and financial-audit implications, and a management presentation with outstanding actions and review status.

\noindent\textbf{P04. Residential Interior Design.} A residential interior-design review coordinates changing room functions, stylistic preferences, furniture, cabinetry, lighting, and equipment interfaces. Successive client discussions and design submissions must be reflected in requirements, decisions, issue registers, and drawing reviews. Delivery assembles a review report, meeting presentation, acceptance documents, and relevant supplied drawings and visual materials. The task is to review and coordinate the evolving design, without independently redrawing all original attachments.

\noindent\textbf{P05. Project Boomerang Forensics.} An audit-committee investigation traces supply-chain fund flows. Fourteen subtasks reconcile bank and business records, check counterparties, prioritize inquiries, revise workpapers, conduct tool-mediated interviews, and integrate additional account evidence into tracing and reproducible calculations. The package contains tracing and calculation workbooks, a methodology note, an editable investigation report, a committee presentation, and handover notes, supported by locally included evidence and unresolved-issue records.

\noindent\textbf{P06. IF Quantitative Research.} An equity-index futures research project assesses readiness for simulated observation. Six subtasks cover data auditing, candidate factors, validation, a frozen modeling plan, execution and account-state simulation, and independent evaluation. Research decisions must connect to implementable trades, reproducible performance calculations, and account constraints. Deliverables include workbooks, code, reports, figures, and a next-stage recommendation; a supported decision not to trade or proceed is permissible.

\noindent\textbf{P07. Qixiang Financial Due Diligence.} Investment-side financial due diligence examines a manufacturing group's consolidation perimeter, financial baseline, earnings quality, and funding needs. Workpapers cover revenue, costs, inventory, working capital, debt, capital expenditure, and personnel. Management responses and supplementary evidence must propagate into adjustments and forecast assessment. Delivery comprises a due diligence report, databook, earnings-adjustment schedule, open-issues register, management presentation, and handover notes linking findings to valuation and transaction decisions.

\noindent\textbf{P08. Pharmacovigilance ICSR Processing.} A post-marketing individual case safety report progresses from receipt through follow-up, medical assessment, quality review, and closure. Later responses and review comments must update clinical facts, reporting classification, narrative, and structured fields without losing source links or unresolved questions. The final narrative, field workbook, closure summary, and archive inventory must reconcile clinical conclusions with review decisions, submission receipts, and administrative status.

\noindent\textbf{P09. Pharmaceutical Commercial Compliance.} A pharmaceutical commercialization-compliance project translates document review and departmental interviews into governance recommendations and policies. Cross-functional feedback must propagate through healthcare-professional interactions, publication review, third-party due diligence, conflicts, investigation cooperation, and conduct rules. The final twelve policy and supporting files, management presentation, and handover recommendations align policy text, approval procedures, forms, and implementation tools while distinguishing existing controls, proposals, unresolved choices, and approved actions.

\noindent\textbf{P10. StarPulse Legal Documentation.} External legal counsel prepares agreements for a platform connecting business clients with professional contributors. Business-model and data-flow review informs the user agreement, privacy policy, and data-submission agreement. Client comments and a cross-border pilot require coordinated revisions across all three documents and launch arrangements. The publishable documents, launch checklist, and handover notes must identify unresolved decisions and release conditions without presenting proposed activities or approvals as established facts.

\noindent\textbf{P11. Evaluation Data Delivery.} An evaluation-data project coordinates acceptance and release across multiple production streams. Task packages, model responses, scoring records, and client feedback undergo registration, deduplication, repair, runtime integration, and targeted testing within a constrained window. Replacements and supplementary deliveries must reconcile across versions. Final outputs include the adopted customer package, evidence, manifests, delivery notes, and internal registers that distinguish production claims, verified results, approvals, dispatch, acceptance, and remaining work.

\FloatBarrier
\Needspace{8\baselineskip}
\section{Evaluation Protocol and Scoring}
\label{app:evaluation}

\subsection{Evaluation Targets and Quality Dimensions}
\label{app:rubric-taxonomy}
Process evaluation assesses subtask submissions using the information available at submission. Final evaluation uses the three targets in Table~\ref{tab:evaluation-targets}. A target specifies the scope of assessment; a quality dimension (Table~\ref{tab:quality-dimensions}) specifies the property checked. Each criterion has one primary quality dimension. The examples illustrate these definitions.

\begin{table}[htbp]
\centering
\small
\caption{Final-delivery evaluation targets and their scopes.}
\label{tab:evaluation-targets}
\begin{tabularx}{\linewidth}{@{}>{\raggedright\arraybackslash}p{0.20\linewidth}>{\raggedright\arraybackslash}X>{\raggedright\arraybackslash}X@{}}
\toprule
Target & Scope & Example criterion \\
\midrule
Artifact quality & Individual deliverables in their submitted final versions. & The final report incorporates the revisions required by the end of the project. \\
Cross-file consistency & Agreement between two or more final artifacts. & The report and workbook agree on the operative assumptions, units, and totals. \\
Package acceptance & The complete file collection against the project's shared delivery requirements. & All required deliverables are included and their handoff status is clearly communicated. \\
\bottomrule
\end{tabularx}
\end{table}

\begin{table}[htbp]
\centering
\small
\caption{Quality dimensions. The same dimension can apply to different evaluation targets.}
\label{tab:quality-dimensions}
\begin{tabularx}{\linewidth}{@{}>{\raggedright\arraybackslash}p{0.20\linewidth}>{\raggedright\arraybackslash}X>{\raggedright\arraybackslash}X@{}}
\toprule
Dimension & Assessed property & Illustrative failure \\
\midrule
Correctness & Accuracy of facts, calculations, reasoning, and conclusions under the applicable information conditions. & A conclusion contradicts the supporting data. \\
Completeness & Coverage of required content, files, and project objectives. & A required analysis or deliverable is missing. \\
Consistency & Agreement of figures, definitions, decisions, and version states within and across artifacts. & A revised assumption is not reflected in an affected report. \\
Professionalism & Adherence to domain methods, terminology, business rules, and responsibility boundaries. & A provisional finding is presented as an authorized final decision. \\
Usability & Organization, presentation, and operational accessibility for the intended recipient. & An otherwise valid workbook cannot be interpreted or used for its intended decision. \\
\bottomrule
\end{tabularx}
\end{table}

\subsection{Criterion Scope, Scoring, and Reuse}
\label{app:score-aggregation}
Criteria specify a task requirement, target artifacts, weight, and judgment evidence. Subtask judgments use the preserved submissions and information conditions; final judgments use the handoff versions and requirements in effect at completion. Artifact-quality assessment reuses applicable source-task criteria while preserving their identities. A criterion or source group contributes once to the final aggregate, even when it covers several files. Previous verdicts are reused only when the relevant artifacts, requirements, and evidence context are unchanged; otherwise the criteria are judged again. Figure~\ref{fig:dataset-overview} counts each unique criterion once.

Final-delivery criteria use signed integer weights with nonzero magnitudes from 1 to 10. Positive weights award credit for satisfied requirements; negative weights deduct credit when the specified failure condition is met. Fatal-gate criteria have weight zero and are checked separately. Dependencies are resolved by the declared criterion rules before aggregation. Each criterion contributes directly to the final score, without an additional weighting layer for evaluation targets or source-task groups. Subtask scores retain their own scoring specifications and are not added to the final score.

Let $w_{p,i}$ be the signed integer weight of criterion $i$, and let $h_{m,p,i}\in\{0,1\}$ be its effective judgment for system $m$ after dependency resolution.
For a positive criterion, $h_{m,p,i}=1$ means the requirement is satisfied; for a negative criterion, it means the specified failure condition is met.
Write $W_p^{+}=\sum_{i\in\mathcal{R}^{\mathrm{final}}_p}\max(w_{p,i},0)>0$, and let $g_{m,p}=1$ indicate that a project-level fatal gate triggers.
The final score is
\begin{equation}
 \mathrm{Final}_{m,p}=\begin{cases}
 0, & g_{m,p}=1,\\[2pt]
 \operatorname{clip}_{[0,100]}\!\left(
 100\,\dfrac{\sum_{i\in\mathcal{R}^{\mathrm{final}}_p}w_{p,i}h_{m,p,i}}{W_p^{+}}
 \right), & g_{m,p}=0.
 \end{cases}
 \label{eq:final-component-score}
\end{equation}
Here $\operatorname{clip}_{[0,100]}$ bounds the normalized score to $[0,100]$.
The same project-specific criterion weights apply to every system.

\subsection{Aggregation and Diagnostic Profiles}
\label{app:reporting-aggregation}
Both reported means give equal weight to the eleven projects:
\begin{equation}
  \overline{\mathrm{Final}}_m = \frac{1}{|\mathcal{P}|}\sum_{p\in\mathcal{P}}\mathrm{Final}_{m,p},
  \qquad
  \overline{\mathrm{Process}}_m = \frac{1}{|\mathcal{P}|}\sum_{p\in\mathcal{P}}\mathrm{Process}_{m,p},
  \label{eq:project-aggregation}
\end{equation}
where $|\mathcal{P}|=11$ and each project's process score first averages its subtask scores (Equation~\ref{eq:delivery-scores}). The subtask pass rate instead counts unrounded subtask scores of at least 90, including subtask gates, over all 150 subtasks per system. Reusing criteria at final delivery does not add subtasks to this denominator. Figure~\ref{fig:process-delivery}(a) reports the difference between process and final means, which assess different artifact versions and requirement sets.

The profiles in Figure~\ref{fig:delivery-diagnostics}(a,b) group final criteria by quality dimension or evaluation target. Within each project and group, signed credit after dependency resolution is divided by the group's positive available weight, scaled to 100, and clipped to $[0,100]$. Group scores are then averaged equally across projects, before project-level fatal gates. Groups without positive available weight are marked unavailable. These separately normalized profiles are diagnostics, not additive parts of the final score; package acceptance is a rubric score rather than an acceptance rate. Missing outcomes remain explicit, and rankings use a shared project set.

\FloatBarrier
\section{Experimental Details}
\label{app:experiments}

\subsection{Agent and Evaluator Configuration}
\label{app:agent-systems}
\label{app:judge-settings}
Table~\ref{tab:agent-configurations} lists the evaluated model--harness pairs and CLI versions. Comparisons concern complete agent systems, including their native tool interfaces and context-management behavior.
\begin{table}[htbp]
\centering
\footnotesize
\caption{Agent-system configurations and CLI versions.}
\label{tab:agent-configurations}
\begin{tabular}{@{}lll@{}}
\toprule
Model & Harness & CLI version \\
\midrule
GPT-5.6 Sol & Codex & 0.142.4 \\
Claude Opus 5 & Claude Code & 2.1.278 \\
Gemini 3.8 Flash & Gemini CLI & 0.60.0 \\
Qwen 3.8 Max & Qwen Code & 0.24.1 \\
DeepSeek V4.1 Flash & Claude Code & 2.1.278 \\
GLM 5.2 & Claude Code & 2.1.278 \\
Kimi K3 & Kimi CLI & 1.50.0 \\
\bottomrule
\end{tabular}
\end{table}

\paragraph{Execution.}
Harbor~0.20.0 orchestrates separate Docker environments for each system--project run under the persistent-workspace protocol in Section~\ref{sec:protocol}. Successive subtask instructions resume the agent's native session rather than create independent conversations, and working files remain available across subtasks. Context management follows each harness's native behavior. New source files are released only when scheduled, while preserved submissions provide immutable evaluation evidence. Each subtask has a 14,400-second agent limit and a separate 1,200-second setup limit. Standard containers request two CPUs, 8~GiB RAM, and 30~GiB storage, with network access enabled. The launcher requests high reasoning effort where supported, without a common token cap or temperature; effective reasoning budgets remain model-dependent.

\paragraph{Evaluation.}
GPT-5.6 Luna runs through Codex CLI~0.142.4, matching the Codex version in Table~\ref{tab:agent-configurations}. It uses high reasoning effort, batches of eight criteria, and up to four concurrent batches. Each batch has at most two attempts with a 1,200-second request timeout, followed by a single-item fallback for unresolved criteria. Retries recover missing responses rather than select favorable judgments. The verifier has a separate 43,200-second limit. The judge assesses evidence; programmatic scoring applies weights, dependencies, normalization, and gates. Evaluator instructions, private references, and verdicts remain outside the agent-facing workspace.

\subsection{Run Provenance and Failure Handling}
\label{app:run-provenance}
The results comprise 77 selected system--project outcomes and 1,050 subtask evaluations, with one reported outcome per pair. Job configurations request one attempt and disable automatic trial retries. Infrastructure recovery can resume preserved subtask submissions or create additional attempt records; these are not independent repetitions. If judging or storage fails after a successful agent submission, recovery preserves that submission instead of regenerating it. Historical scores from different configurations are not pooled without reassessment.

Missing deliverables in completed submissions are evaluated by the applicable criteria and gates. Pending runs, infrastructure interruptions, and missing judgments retain explicit status; completed evaluations with triggered fatal conditions retain their zero final scores. For the evaluated GLM system, Claude Code's native PDF and image payloads are unsupported: a compatibility guard returns recoverable tool errors while keeping local files and extraction tools available. This interface adaptation is recorded with the run provenance.

\subsection{Resource Accounting}
\label{app:resource-accounting}
Figure~\ref{fig:delivery-diagnostics}(c) sums agent execution intervals within each project, including model waiting and tool execution but excluding setup and judging. Recovered submissions use their original intervals. All 1,050 selected subtasks are covered: 1,047 use host-recorded intervals, two native CLI durations, and one an interval inferred from native session timestamps.

Token totals are available for all 77 outcomes and deduplicated from retained native sessions, including identifiable logged subagents. Total tokens equal input plus output; cached input is already included in input and is not added again. Unavailable usage fields remain missing. Reported resources cover the selected scored subtasks and retained sessions, excluding additional overhead from historical failed attempts and recovery. Tokenization, reasoning-token reporting, and caching differ by provider, so token counts measure usage rather than comparable monetary cost.

\clearpage
\section{Complete Project Results}
\label{app:results}
\subsection{Agent--Project Scores}
\label{app:project-scores}
Table~\ref{tab:project-scores} reports all 77 system--project pairs in two panels. Scores range from 0 to 100, rounded to two decimals only for display. Columns correspond to the model--harness systems in Table~\ref{tab:agent-configurations}; project identifiers follow Appendix~\ref{app:project-profiles}. Final delivery and process evaluation assess the distinct targets specified in Section~\ref{sec:protocol}.

\begin{table}[!htbp]
\centering\small
\setlength{\tabcolsep}{4pt}
\renewcommand{\arraystretch}{1.10}
\caption{Final-delivery and mean process scores for all seven agent systems and eleven projects. Final scores retain acceptance gates; process scores average subtask submissions within each project. Mean rows weight projects equally.}
\label{tab:project-scores}
\label{tab:project-final-scores}
\label{tab:project-process-scores}
\begin{tabularx}{\linewidth}{@{}l*{7}{>{\centering\arraybackslash}X}@{}}
\toprule
\multicolumn{8}{@{}l}{\textbf{Final delivery}} \\
\addlinespace[2pt]
Project & Claude & GPT & Qwen & DeepSeek & Kimi & GLM & Gemini \\
\midrule
P01 & 58.60 & 54.00 & 47.94 & 47.22 & 58.84 & 50.36 & 26.63 \\
P02 & 40.57 & 28.37 & 34.83 & 41.83 & 0.00 & 40.22 & 37.52 \\
P03 & 58.19 & 64.92 & 72.27 & 71.22 & 73.11 & 60.08 & 35.08 \\
P04 & 47.01 & 67.66 & 61.41 & 60.87 & 61.14 & 54.08 & 45.65 \\
P05 & 84.15 & 71.47 & 60.52 & 63.11 & 70.32 & 63.40 & 40.63 \\
P06 & 78.44 & 75.33 & 78.44 & 63.11 & 72.67 & 52.67 & 44.44 \\
P07 & 54.59 & 49.38 & 56.58 & 48.88 & 58.81 & 33.25 & 46.90 \\
P08 & 66.37 & 59.65 & 55.26 & 61.70 & 71.64 & 0.00 & 47.08 \\
P09 & 75.27 & 76.36 & 66.91 & 56.73 & 54.18 & 50.18 & 54.18 \\
P10 & 69.57 & 77.81 & 55.31 & 52.46 & 45.32 & 32.33 & 41.84 \\
P11 & 82.20 & 76.27 & 74.86 & 72.88 & 87.29 & 52.54 & 24.29 \\
\midrule
\textbf{Mean} & \textbf{65.00} & \textbf{63.75} & \textbf{60.39} & \textbf{58.18} & \textbf{59.39} & \textbf{44.46} & \textbf{40.39} \\
\midrule
\addlinespace[3pt]
\multicolumn{8}{@{}l}{\textbf{Process evaluation}} \\
\addlinespace[2pt]
Project & Claude & GPT & Qwen & DeepSeek & Kimi & GLM & Gemini \\
\midrule
P01 & 87.33 & 86.17 & 78.39 & 85.61 & 82.50 & 78.17 & 63.00 \\
P02 & 50.69 & 35.15 & 44.54 & 43.00 & 44.31 & 37.46 & 44.15 \\
P03 & 82.00 & 78.88 & 84.00 & 80.44 & 80.75 & 76.81 & 64.19 \\
P04 & 76.60 & 86.20 & 82.20 & 87.20 & 82.80 & 81.50 & 80.80 \\
P05 & 86.36 & 79.93 & 73.57 & 80.71 & 74.93 & 75.14 & 56.86 \\
P06 & 75.67 & 75.54 & 70.83 & 62.33 & 62.25 & 55.71 & 44.88 \\
P07 & 88.42 & 83.39 & 80.30 & 78.42 & 79.27 & 57.52 & 63.82 \\
P08 & 79.83 & 72.92 & 65.83 & 68.50 & 61.00 & 51.42 & 50.58 \\
P09 & 85.63 & 79.71 & 71.13 & 78.18 & 72.56 & 59.63 & 54.99 \\
P10 & 72.56 & 75.22 & 66.12 & 62.28 & 55.38 & 43.16 & 49.88 \\
P11 & 92.20 & 91.07 & 88.53 & 86.67 & 81.47 & 76.53 & 65.93 \\
\midrule
\textbf{Mean} & \textbf{79.75} & \textbf{76.74} & \textbf{73.22} & \textbf{73.94} & \textbf{70.66} & \textbf{63.00} & \textbf{58.10} \\
\bottomrule
\end{tabularx}
\end{table}

\subsection{Resource Use}
\label{app:resource-use}
Tables~\ref{tab:resource-summary} and~\ref{tab:project-resources} report the values underlying Figure~\ref{fig:delivery-diagnostics}(c). Time sums agent execution intervals within each project, including model waiting and tool execution but excluding setup and judging. Total tokens are retained native-session input plus output, with cached input included once. Values follow the accounting in Appendix~\ref{app:resource-accounting}; means weight the eleven projects equally, and rounding is applied only for display.

\begin{table}[htbp]
\centering\small
\caption{Mean execution time and total tokens per project for the evaluated agent systems.}
\label{tab:resource-summary}
\begin{tabular}{@{}lrr@{}}
\toprule
System & Time (hours) & Tokens (millions) \\
\midrule
Claude Opus 5 & 10.51 & 124.12 \\
GPT-5.6 Sol & 2.05 & 36.43 \\
Qwen 3.8 Max & 4.92 & 112.90 \\
DeepSeek V4.1 Flash & 8.18 & 184.47 \\
Kimi K3 & 7.92 & 27.06 \\
GLM 5.2 & 2.60 & 25.97 \\
Gemini 3.8 Flash & 2.25 & 85.65 \\
\bottomrule
\end{tabular}
\end{table}

\begin{table}[htbp]
\centering\small
\setlength{\tabcolsep}{4pt}
\renewcommand{\arraystretch}{1.08}
\caption{Execution time and total tokens for all 77 system--project outcomes. System columns follow Table~\ref{tab:agent-configurations}; project names are given in Table~\ref{tab:project-names}.}
\label{tab:project-resources}
\begin{tabularx}{\linewidth}{@{}l*{7}{>{\centering\arraybackslash}X}@{}}
\toprule
\multicolumn{8}{@{}l}{\textbf{Execution time (hours)}} \\
Project & Claude & GPT & Qwen & DeepSeek & Kimi & GLM & Gemini \\
\midrule
P01 & 13.91 & 2.43 & 4.87 & 9.89 & 6.71 & 4.08 & 2.65 \\
P02 & 10.74 & 1.64 & 4.57 & 5.62 & 7.42 & 2.90 & 1.76 \\
P03 & 12.36 & 2.08 & 2.93 & 5.05 & 3.89 & 2.17 & 2.05 \\
P04 & 9.75 & 1.79 & 2.72 & 5.79 & 6.35 & 1.84 & 2.16 \\
P05 & 12.03 & 2.87 & 4.46 & 12.74 & 8.64 & 4.78 & 2.04 \\
P06 & 7.44 & 0.65 & 8.72 & 2.52 & 6.26 & 1.69 & 1.30 \\
P07 & 26.71 & 5.12 & 7.85 & 32.22 & 21.72 & 5.43 & 5.54 \\
P08 & 2.14 & 0.86 & 1.82 & 1.96 & 2.52 & 0.96 & 1.97 \\
P09 & 11.98 & 2.84 & 2.88 & 7.81 & 18.52 & 2.67 & 2.48 \\
P10 & 2.92 & 0.88 & 2.57 & 1.52 & 1.73 & 0.76 & 1.41 \\
P11 & 5.66 & 1.44 & 10.76 & 4.89 & 3.38 & 1.26 & 1.35 \\
\midrule
\textbf{Mean} & \textbf{10.51} & \textbf{2.05} & \textbf{4.92} & \textbf{8.18} & \textbf{7.92} & \textbf{2.60} & \textbf{2.25} \\
\midrule
\addlinespace[3pt]
\multicolumn{8}{@{}l}{\textbf{Total tokens (millions)}} \\
Project & Claude & GPT & Qwen & DeepSeek & Kimi & GLM & Gemini \\
\midrule
P01 & 171.59 & 46.62 & 71.03 & 209.63 & 19.84 & 31.18 & 116.06 \\
P02 & 176.86 & 33.48 & 142.40 & 334.39 & 26.44 & 29.68 & 80.69 \\
P03 & 173.83 & 39.83 & 78.88 & 121.98 & 16.47 & 22.20 & 95.48 \\
P04 & 79.67 & 32.19 & 18.83 & 92.56 & 20.69 & 18.09 & 74.58 \\
P05 & 149.70 & 50.90 & 130.75 & 368.70 & 37.14 & 55.08 & 110.69 \\
P06 & 107.92 & 12.47 & 253.88 & 71.17 & 20.87 & 23.40 & 51.76 \\
P07 & 251.85 & 80.08 & 189.77 & 560.04 & 64.31 & 46.66 & 190.11 \\
P08 & 15.18 & 16.33 & 30.05 & 22.54 & 8.62 & 6.78 & 48.25 \\
P09 & 147.56 & 45.20 & 49.62 & 162.71 & 66.82 & 21.91 & 64.79 \\
P10 & 28.56 & 18.60 & 48.28 & 24.18 & 5.57 & 7.96 & 46.69 \\
P11 & 62.62 & 25.05 & 228.47 & 61.26 & 10.87 & 22.79 & 63.10 \\
\midrule
\textbf{Mean} & \textbf{124.12} & \textbf{36.43} & \textbf{112.90} & \textbf{184.47} & \textbf{27.06} & \textbf{25.97} & \textbf{85.65} \\
\bottomrule
\end{tabularx}
\end{table}

\FloatBarrier

\subsection{Project Trajectories}
\label{app:project-trajectories}
\begin{figure}[!htbp]
  \centering
  \includegraphics[width=\linewidth]{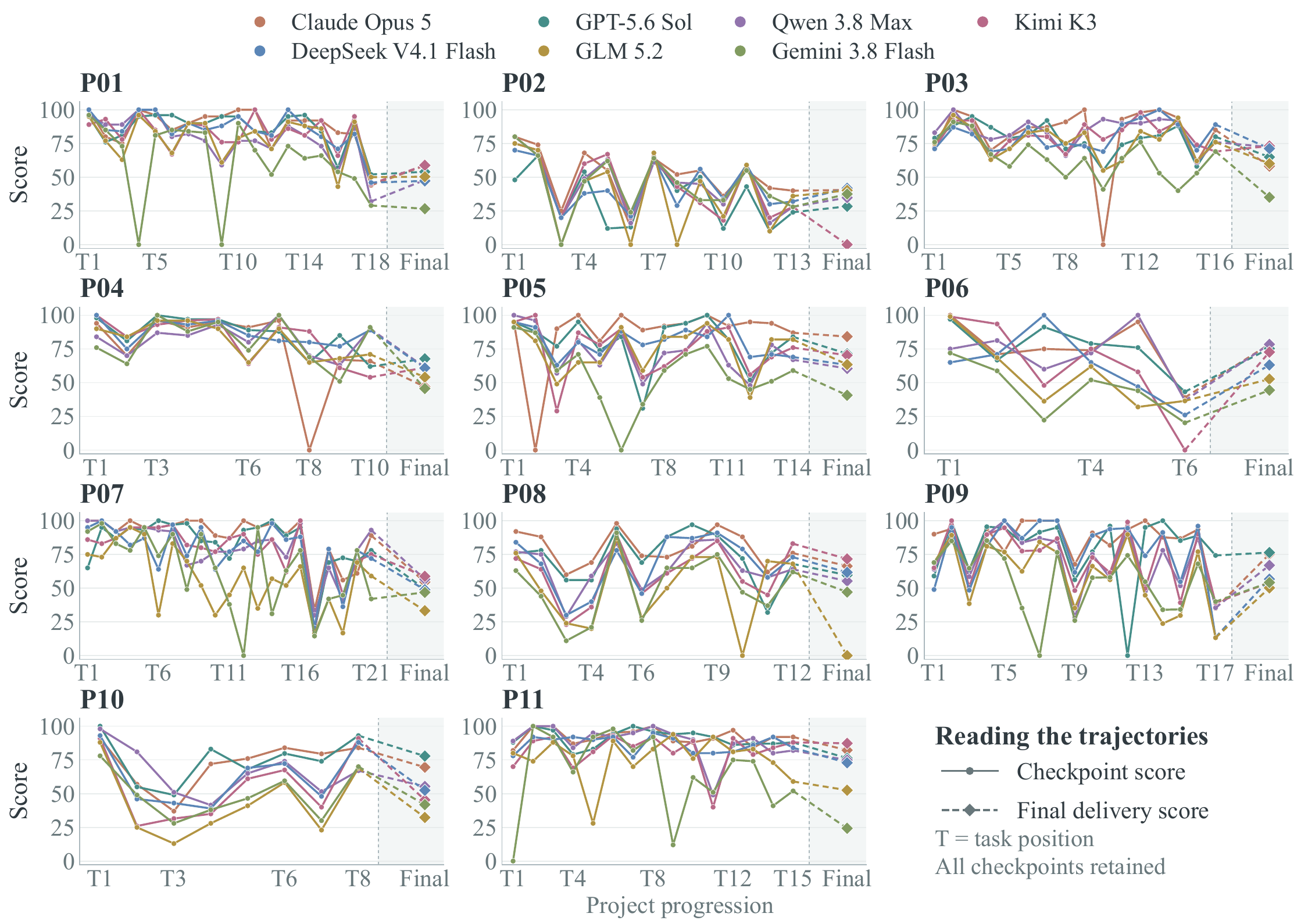}
  \caption{\textbf{Agent trajectories across all 11 projects.} Panels are ordered P01--P11. Circles show subtask scores; diamonds show final-delivery scores. T labels denote chronological subtask positions, with all original subtask submissions retained. Dashed terminal segments mark the change in evaluation scope. All panels share the same score scale and agent colors.}
  \label{fig:all-project-trajectories}
\end{figure}

Figure~\ref{fig:all-project-trajectories} extends the three cases in Section~\ref{sec:case-trajectories} to all 11 projects and seven agents, retaining all 1,050 subtask scores and 77 final-delivery scores.
The panels use the same score scale and agent colors, allowing the quality changes discussed in the main text to be examined across the full benchmark.

\FloatBarrier

\end{document}